\documentclass[twoside,11pt]{article}

\usepackage[table]{xcolor}
\usepackage{graphicx} 
\usepackage{subfig}
\usepackage{algorithm}
\usepackage{algorithmic}
\usepackage{amsmath}
\usepackage{amssymb}
\usepackage{amsfonts}
\usepackage{dsfont}
\usepackage{xspace}
\usepackage{xr}
\usepackage{multirow}
\usepackage{multicol}
\usepackage{siunitx}
\usepackage{afterpage}
\usepackage{enumerate}
\usepackage{booktabs}
\usepackage{natbib}
\usepackage{soul}
\usepackage{mathtools}

\usepackage{jmlr2e}

\def\x{{\mathbf x}}

\def\S{{\cal S}}
\def\T{{\cal T}}

\def\R{{\mathds R}}

\def\tf{{\theta_f}}
\def\td{{\theta_d}}
\def\ty{{\theta_y}}
\def\htf{{\hat\theta_f}}
\def\htd{{\hat\theta_d}}
\def\hty{{\hat\theta_y}}

\usepackage{tikz}
\usetikzlibrary{positioning}
\usetikzlibrary{calc}

\usepackage{pgfplots}
\pgfplotsset{compat=newest} 
\pgfplotsset{plot coordinates/math parser=false}

\newlength\figureheight
\newlength\figurewidth

\newcommand{\fig}[1]{Figure~\ref{fig:#1}}
\newcommand{\sect}[1]{Section~\ref{sect:#1}}
\newcommand{\tab}[1]{Table~\ref{tab:#1}}
\newcommand{\eq}[1]{(\ref{eq:#1})}

\makeatletter
\newcommand{\todo}[1][]{\@latex@warning{TODO #1}\fbox{TODO\dots}}
\makeatother

\usepackage{quebec-def}

\newcommand{\uu}{{\mathbf{u}}}

\newcommand{\redplus}{``\red{$\boldsymbol{{+}}$}''}
\newcommand{\greenminus}{``\green{$\boldsymbol{{\pmb{-}}}$}''}

\renewcommand{\eqdef}{=}
\renewcommand{\Xcal}{X}
\renewcommand{\Ycal}{Y}
\renewcommand{\Rbb}{\R}

\renewcommand{\S}{\DSX}
\renewcommand{\T}{\DTX}

\renewcommand{\xbs}{\xb}
\renewcommand{\xbt}{\xb}
\newcommand{\ys}{y}
\newcommand{\yt}{y}

\begin{document} 

\title{Domain-Adversarial Training of Neural Networks}

\jmlrheading{17}{2016}{1-35}{5/15}{4/16}{Yaroslav Ganin, Evgeniya Ustinova, Hana Ajakan, Pascal Germain, Hugo Larochelle, et al}

\ShortHeadings{Domain-Adversarial Neural Networks}{Ganin, Ustinova, Ajakan, Germain, Larochelle, Laviolette, Marchand and Lempitsky} 

\author{%
\name Yaroslav Ganin \email ganin@skoltech.ru\\
\name Evgeniya Ustinova \email evgeniya.ustinova@skoltech.ru\\
	\addr 
Skolkovo Institute of Science and Technology (Skoltech)\\
Skolkovo, Moscow Region, Russia	
\AND
\name Hana Ajakan \email hana.ajakan.1@ulaval.ca\\
\name Pascal Germain   \email Pascal.Germain@ift.ulaval.ca \\
	\addr D{\'e}partement d'informatique et de g\'enie logiciel,
	Universit\'e Laval  \\
	Qu\'ebec, Canada,
	G1V 0A6
\AND
\name Hugo Larochelle    \email hugo.larochelle@usherbrooke.ca \\
	\addr D{\'e}partement d'informatique, 
	Universit\'e de Sherbrooke  \\
	Qu\'ebec, Canada,
	J1K 2R1 
\AND
\name Fran\c{c}ois Laviolette    \email Francois.Laviolette@ift.ulaval.ca \\
\name Mario Marchand    \email Mario.Marchand@ift.ulaval.ca \\
	\addr D{\'e}partement d'informatique et de g\'enie logiciel, 
	Universit\'e Laval  \\
	Qu\'ebec, Canada,
	G1V 0A6    
\AND
\name Victor Lempitsky    \email lempitsky@skoltech.ru \\
	\addr 
Skolkovo Institute of Science and Technology (Skoltech)\\
Skolkovo, Moscow Region, Russia	
}

\editor{Urun Dogan, Marius Kloft, Francesco Orabona, and Tatiana Tommasi}

\maketitle

\begin{abstract}%
We introduce a new representation learning approach for domain adaptation, in which data at training and test time
come from similar but different distributions. Our approach is directly
inspired by the theory on domain adaptation suggesting that, for effective domain
transfer to be achieved, predictions must be made based on features that cannot discriminate between the training (source)
and test (target) domains. 

The approach implements this idea in the context of neural network architectures that are trained on labeled data from the source domain and unlabeled data from the target domain (no labeled target-domain data is necessary). As the training progresses, the approach promotes the emergence of features that are (i) discriminative for the main learning task on the source domain and (ii) indiscriminate with respect to the shift between the domains. We show that this adaptation behaviour can be achieved in almost any feed-forward model by augmenting it with few standard layers and a new {\em gradient reversal} layer. The resulting augmented architecture can be trained using standard backpropagation and stochastic gradient descent, and can thus be implemented with little effort using any of the deep learning packages. 

We demonstrate the success of our approach for two distinct classification problems (document sentiment analysis and image classification), where state-of-the-art domain adaptation performance on standard benchmarks is achieved. We also validate the approach for descriptor learning task in the context of person re-identification application.
\end{abstract} 


\section{Introduction}

The cost of generating labeled data for a new machine learning task is
often an obstacle for applying machine learning methods. 
In particular, this is a limiting factor for the further progress of
deep neural network architectures, that have already brought impressive advances to the state-of-the-art across a wide variety of machine-learning tasks and applications. 
For problems lacking labeled data, it may be still possible to obtain training sets that are big enough for training large-scale deep models, but that suffer from the {\em shift} in data distribution from the actual data encountered at ``test time''. One  important example is training an image classifier on synthetic or semi-synthetic images, which may come in abundance and be fully labeled, but which inevitably have a distribution that is different from real images~\citep{Liebelt10,Stark10,Vazquez14,Sun14}.  Another example is in the context of sentiment analysis in written reviews, where one  might have labeled data for reviews of one type of product (\eg, movies), while having the need to classify reviews of other products (\eg, books).

Learning a discriminative classifier or other predictor in the presence of a {\em shift} between training and test distributions is known as {\em domain adaptation} (DA).
The proposed approaches build mappings between the {\em source} (training-time) and the {\em target} (test-time) domains, so that the classifier learned for the source domain can also be applied to the target domain, when composed with the learned mapping between domains. The appeal of the domain adaptation approaches is the ability to learn a mapping between domains in the situation when the target domain data are either fully unlabeled ({\em unsupervised domain annotation}) or have few labeled samples ({\em semi-supervised domain adaptation}). Below, we focus on the harder unsupervised case, although the proposed approach (\textit{domain-adversarial learning}) can be generalized to the semi-supervised case rather straightforwardly.

Unlike many previous papers on domain adaptation that worked with fixed feature representations, we focus on combining domain adaptation and deep feature learning within one training process. %
Our goal is to embed domain adaptation into the process of learning representation, so that the final classification decisions are made based on features that are both discriminative and invariant to the change of domains, \ie, have the same or very similar distributions in the source and the target domains. In this way, the obtained feed-forward network can be applicable to the target domain without being hindered by the shift between the two domains.  Our approach is motivated by the theory on domain
adaptation~\citep{BenDavid-NIPS06,BenDavid-MLJ2010}, that suggests that
a good representation for cross-domain transfer is one for which an algorithm cannot learn to identify the domain of origin of the input observation.

We thus focus on learning features that combine (i) discriminativeness and (ii) domain-invariance. This is achieved by jointly optimizing the underlying features as well as two discriminative classifiers operating on these features: (i) the \emph{label predictor} that predicts class labels and is used both during training and at test time and (ii) the \emph{domain classifier} that discriminates between the source and the target domains during training. While the parameters of the classifiers are optimized in order to minimize their error on the training set, the parameters of the underlying deep feature mapping are optimized in order to {\em minimize} the loss of the label classifier and to {\em maximize} the loss of the domain classifier. The latter update thus works \textit{adversarially} to the domain classifier, and it encourages domain-invariant features to emerge in the course of the optimization.

Crucially, we show that all three training processes can be embedded into an appropriately composed deep feed-forward network, called \textit{domain-adversarial neural network} (DANN) (illustrated by \fig{arch}, page~\pageref{fig:arch}) that uses standard layers and loss functions, and can be trained using standard backpropagation algorithms based on stochastic gradient descent or its modifications (\eg, SGD with momentum). The approach is generic as a DANN version can be created for almost any existing feed-forward architecture that is trainable by backpropagation. In practice, the only non-standard component of the proposed architecture is a rather trivial {\em gradient reversal} layer that leaves the input unchanged during forward propagation and reverses the gradient by multiplying it by a negative scalar during the backpropagation.

We provide an experimental evaluation of the proposed domain-adversarial learning idea over a range of deep architectures and applications. We first consider the simplest DANN architecture where the three parts (label predictor, domain classifier and feature extractor) are linear, and demonstrate the success of domain-adversarial learning for such architecture. The evaluation is performed for synthetic data as well as for the sentiment analysis problem in natural language processing, where DANN improves the state-of-the-art \emph{marginalized Stacked Autoencoders} (mSDA) of~\citet{Chen12} on the common Amazon reviews benchmark.

We further evaluate the approach extensively for an image classification task, and present results on traditional deep learning image data sets---such as MNIST~\citep{LeCun98} and SVHN~\citep{Netzer11}---as well as on {\sc Office} benchmarks \citep{Saenko10}, where  domain-adversarial learning allows obtaining a deep architecture that considerably improves over previous state-of-the-art accuracy. 

Finally, we evaluate domain-adversarial \textit{descriptor} learning in the context of person re-identification application~\citep{gong2014person}, where the task is to obtain good pedestrian image descriptors that are suitable for retrieval and verification. We apply domain-adversarial learning, as we consider a \textit{descriptor predictor} trained with a Siamese-like loss instead of the label predictor trained with a classification loss. In a series of experiments, we demonstrate that domain-adversarial learning can improve cross-data-set re-identification considerably.

\section{Related work}
\label{sec:related}

The general approach of achieving domain adaptation
explored under many facets. Over the years, a large part of the literature has focused mainly on linear 
hypothesis \citep[see for instance][]{BlitzerMP06,BruzzoneM10S,pbda,Baktashmotlagh13,CortesM14}.
More recently, non-linear representations have become increasingly studied,
including neural network representations~\citep{Glorot11,LiY2014} and most notably
the state-of-the-art mSDA~\citep{Chen12}. That literature has mostly focused
on exploiting the principle of robust representations, based on the denoising autoencoder paradigm~\citep{VincentP2008}. 

Concurrently, multiple methods of matching the feature distributions in the source and the target domains have been proposed for unsupervised domain adaptation. Some approaches perform this by reweighing or selecting samples from the source domain \citep{Borgwardt06,Huang06,Gong13}, while others seek an explicit feature space transformation that would map source distribution into the target one \citep{Pan11,Gopalan11,Baktashmotlagh13}. An important aspect of the distribution matching approach is the way the (dis)similarity between distributions is measured.  Here, one popular choice is matching the distribution means in the kernel-reproducing Hilbert space \citep{Borgwardt06,Huang06}, whereas \citet{Gong12} and \citet{Fernando13} map the principal axes associated with each of the distributions. 

Our approach also attempts to match feature space distributions, however this is accomplished by modifying the feature representation itself rather than by reweighing or geometric transformation.
Also, our method uses a rather different way to measure the disparity between distributions based on their separability by a deep discriminatively-trained classifier. 
Note also that several approaches perform transition from the source to the target domain \citep{Gopalan11,Gong12} by changing gradually the training distribution. Among these methods, \citet{Chopra13} does this in a ``deep'' way by the layerwise training of a sequence of deep autoencoders, while gradually replacing source-domain samples with target-domain samples. This improves over a similar approach of \citet{Glorot11} that simply trains a single deep autoencoder for both domains. In both approaches, the actual classifier/predictor is learned in a separate step using the feature representation learned by autoencoder(s). In contrast to \citet{Glorot11,Chopra13}, our approach performs feature learning, domain adaptation and classifier learning jointly, in a unified architecture, and using a single learning algorithm (backpropagation). We therefore argue that our approach is simpler (both conceptually and in terms of its implementation). Our method also achieves considerably better results on the popular {\sc Office} benchmark.

While the above approaches perform unsupervised domain adaptation, there are approaches that perform {\em supervised} domain adaptation by exploiting labeled data from the target domain. In the context of deep feed-forward architectures, such data can be used to ``fine-tune'' the network trained on the source domain~\citep{Zeiler13,Oquab14,Babenko14}. Our approach does not require labeled target-domain data. At the same time, it can easily incorporate such data when they are available.

An idea related to ours is described in \citet{Goodfellow14}. While their goal is quite different (building generative deep networks that can synthesize samples), the way they measure and minimize the discrepancy between the distribution of the training data and the distribution of the synthesized data is very similar to the way our architecture measures and minimizes the discrepancy between feature distributions for the two domains. Moreover, the authors mention the problem  of saturating sigmoids which may arise at the early stages of training due to the significant dissimilarity of the domains. The technique they use to circumvent this issue (the ``adversarial'' part of the gradient is replaced by a gradient computed with respect to a suitable cost) is directly applicable to our method. 

Also, recent and concurrent reports by \citet{Tzeng14,Long15} focus on domain adaptation in feed-forward networks. Their set of techniques measures and minimizes the distance between the data distribution means across domains (potentially, after embedding distributions into RKHS). Their approach is thus different from our idea of matching distributions by making them indistinguishable for a discriminative classifier. Below, we compare our approach to \citet{Tzeng14,Long15} on the Office benchmark. Another approach to deep domain adaptation, which is arguably more different from ours, has been developed in parallel by \citet{Chen15}.

From a theoretical standpoint, our approach is directly derived from the seminal theoretical works of \citet{BenDavid-NIPS06,BenDavid-MLJ2010}. Indeed, DANN directly optimizes the notion of $\Hcal$-divergence. We do note the work of \citet{HuangY12},
in which HMM representations are learned for word tagging using a posterior regularizer that
is also inspired by \citeauthor{BenDavid-MLJ2010}'s work. In addition to the tasks being different---\citet{HuangY12} focus on word tagging problems---, we would argue that DANN learning objective more closely optimizes
the $\Hcal$-divergence, with \citet{HuangY12} relying on cruder approximations 
for efficiency reasons.

A part of this paper has been published as a conference paper \citep{moscow}. This version extends \citet{moscow} very considerably by incorporating the report \citet{quebec} (presented as part of the \textit{Second Workshop on Transfer and Multi-Task Learning}), which brings in new terminology, in-depth theoretical analysis and justification of the approach, extensive experiments with the shallow DANN case on synthetic data as well as on a natural language processing task (sentiment analysis). Furthermore, in this version we go beyond classification and evaluate domain-adversarial learning for descriptor learning setting within the person re-identification application. 

\section{Domain Adaptation}
\label{section:DA_theory}

We consider classification tasks where $\Xcal$ is the input space and $\Ycal=\{0,1,\ldots,L{-}1\}$ is the set of $L$ possible labels.
Moreover, we have two different distributions over $\Xcal\!\times\! \Ycal$, called the {\it source domain} $\DS$ and the {\it target domain} $\DT$.
An \emph{unsupervised domain adaptation} learning algorithm is then provided with a {\it labeled source sample} $S$ drawn {\it i.i.d.} from $\DS$, and an {\it unlabeled target sample} $T$ drawn {\it i.i.d.} from $\DTX$, where $\DTX$ is the marginal distribution of $\DT$ over~$\Xcal$.
\begin{equation*}
S = \{(\xbs_i,\ys_i)\}_{i=1}^{n} \sim (\DS)^n 
\,; \quad
T = \{\xbt_i\}_{i=n+1}^{N} \sim (\DTX)^{n'},
\end{equation*}
with $N=n+n'$ being the total number of samples. 
The goal of the learning algorithm is to build a classifier $\eta:\Xcal\to\Ycal$ with a low \emph{target~risk}
\begin{equation*}
\RDT(\eta) \ \eqdef \Pr_{(\xbt,\yt) \sim \DT} \Big(\eta(\xbt) \neq \yt\Big)\,,
\end{equation*}
while having no information about the labels of $\DT$.

\subsection{Domain Divergence}

To tackle the challenging domain adaptation task, many  approaches bound the target error by the sum of the source error and a notion of distance between the source and the target distributions. These methods are intuitively justified by a simple assumption: the source risk is expected to be a good indicator of the target risk when both distributions are similar. Several notions of distance have been proposed for domain adaptation \citep{BenDavid-NIPS06,BenDavid-MLJ2010,Mansour-COLT09,MansourMR09,pbda}.
In this paper, we focus on the $\Hcal$-divergence used by \citet{BenDavid-NIPS06,BenDavid-MLJ2010}, and based on the earlier work of \citet{kifer-2004}. Note that we assume in definition~\ref{def:Hdiv} below that the hypothesis class $\Hcal$ is a (discrete or continuous) set of binary classifiers $\eta:\Xcal\to\{0,1\}$.\footnote{As mentioned by \citet{BenDavid-NIPS06}, the same analysis holds for multiclass setting. However, to obtain the same results when $|Y|>2$, one should assume that $\Hcal$ is a symmetrical hypothesis class. That is, for all $h\in\Hcal$ and any permutation of labels $c:Y\to Y$, we have $c(h)\in \Hcal$. Note that this is the case for most commonly used neural network architectures.}
\begin{definition}[\citealp{BenDavid-NIPS06,BenDavid-MLJ2010, kifer-2004}] \label{def:Hdiv}
Given two domain distributions $\DSX$ and $\DTX$ over~$\Xcal$, and a hypothesis class~$\Hcal$, the \emph{$\Hcal$-divergence} between $\DSX$ and $\DTX$ is
\begin{eqnarray*}
d_\Hcal(\DSX,\DTX)  &\eqdef & 
2 \,\sup_{\eta\in\Hcal} \,\bigg|\, 
\Pr_{\xbs \sim \DSX} \big[\eta(\xbs) = 1\big] - 
\Pr_{\xbt \sim \DTX} \big[\eta(\xbt) = 1\big]\,
\bigg|\,.
\end{eqnarray*}
\end{definition}

That is, the $\Hcal$-divergence relies on the capacity of the hypothesis class $\Hcal$ to distinguish between examples generated by $\DSX$ from examples generated by $\DTX$.
\citet{BenDavid-NIPS06,BenDavid-MLJ2010} proved that, for a symmetric hypothesis class $\Hcal$, one can compute the \emph{empirical $\Hcal$-divergence} between two samples $S\sim(\DSX)^n$ and  $T\sim(\DTX)^{n'}$ by computing
\begin{eqnarray} \label{eq:Hdiv_empirique}
 \hat{d}_\Hcal(S,T) & \eqdef & 
 2\,\Bigg( 1 - \min_{\eta\in\Hcal} \bigg[
\frac{1}{n} \sum_{i=1}^n I[\eta(\xbs_i)\!=\!0] + \frac{1}{n'} \sum_{i=n+1}^{N} I[\eta(\xbt_i)\!=\!1]
\bigg] \Bigg)\,,
\end{eqnarray}
where $I[a]$ is the indicator function which is $1$ if predicate $a$ is true, and $0$ otherwise.

\subsection{Proxy Distance}
\label{section:PAD}

\citet{BenDavid-NIPS06} suggested that, even if it is generally hard to compute $\hat{d}_\Hcal(S,T)$ exactly (\eg, when $\Hcal$ is the space of linear classifiers on $\Xcal$), we can easily approximate it by running a learning algorithm on the problem of discriminating between source and target examples. To do so, we construct a new data set
\begin{equation}\label{eq:U}
U \ =\ \{(\xbs_i, 0)\}_{i=1}^n \cup \{(\xbt_i, 1)\}_{i=n+1}^{N}\,,
\end{equation}
where the examples of the source sample are labeled $0$ and the examples of the target sample are labeled $1$. Then, the risk of the classifier trained on the new data set $U$ approximates the ``$\min$'' part of Equation~\eqref{eq:Hdiv_empirique}. 
Given a generalization error~$\epsilon$ on the problem of discriminating between source and target examples, the $\Hcal$-divergence is then approximated by
\begin{equation} \label{eq:PAD}
\hat{d}_\Acal \ = \ 2\,(1-2\epsilon)\,.
\end{equation}
In \citet{BenDavid-NIPS06}, the value $\hat{d}_\Acal$ is called the \emph{Proxy $\Acal$-distance} (PAD). The \emph{$\Acal$-distance}
being defined as 
$
d_\Acal(\DSX,\DTX)  \eqdef 
2 \,\sup_{A\in\Acal} \,\big|\, 
\Pr_{\DSX} (A) - 
\Pr_{\DTX} (A)\,
\big|
$,
where $\Acal$ is a subset of $\Xcal$. Note that, by choosing $\Acal = \{A_\eta | \eta \in \Hcal\}$, with $A_\eta$ the set represented by the characteristic function $\eta$, the $\Acal$-distance and the $\Hcal$-divergence of Definition~\ref{def:Hdiv} are identical.

In the experiments section of this paper, we compute the PAD value following the approach of \citet{Glorot11,Chen12}, \ie, we train either a linear SVM or a deeper MLP classifier on a subset of $U$ (Equation~\ref{eq:U}), and we use the obtained classifier error on the other subset as the value of~$\epsilon$ in Equation~\eqref{eq:PAD}. 
More details and illustrations of the linear SVM case are provided in Section~\ref{section:PAD_experiments}.

\subsection{Generalization Bound on the Target Risk}
The work of \citet{BenDavid-NIPS06,BenDavid-MLJ2010} also showed that the $\Hcal$-divergence $d_\Hcal(\DSX,\DTX)$ is upper bounded by its empirical estimate $\hat{d}_\Hcal(S,T)$ plus a constant complexity term that depends on the \emph{VC dimension} of $\Hcal$ and the size of samples $S$ and $T$. By combining this result with a similar bound on the source risk, the following theorem is obtained.
\begin{theorem}[\citealp{BenDavid-NIPS06}] 
\label{thm:RDT_bound}
Let $\Hcal$ be a hypothesis class of VC dimension $d$.
With probability $1-\delta$ over the choice of samples $S\sim (\DS)^n$ and $T\sim (\DTX)^{n}$, for every $\eta\in\Hcal$:
\begin{eqnarray*}
\RDT(\eta) &\leq&  
\RS(\eta) +  \sqrt{\frac{4}{n}\left( d \log\tfrac{2e\, n}{d}+  \log\tfrac{4}{\delta}\right) } 
+\hat{d}_\Hcal(S,T) + 4\sqrt{ \frac{1}{n}\left(d \log\tfrac{2 n}{d}+  \log\tfrac{4}{\delta}\right) }
+ \beta\,,
\end{eqnarray*}
with $\beta \geq {\displaystyle\inf_{\eta^*\in\Hcal}} \left[ \RDS(\eta^*) + \RDT(\eta^*) \right]$\,, 
and 
«\begin{equation*}
\RS(\eta) \ =\ \frac{1}{n}\dsum_{i=1}^m I\left[\eta(\xbs_i) \neq \ys_i\right]
\end{equation*}
is the {empirical source~risk}.
\end{theorem}
The previous result tells us that $\RDT(\eta)$ can be low only when the $\beta$ term is low, \ie, only when there exists a classifier that can achieve a low risk on both distributions. It also tells us that, to find a classifier with a small $\RDT(\eta)$ in a given class of fixed VC dimension, the learning algorithm should minimize (in that class) a trade-off between the source risk $\RS(\eta)$ and the empirical $\Hcal$-divergence $\hat{d}_\Hcal(S,T)$.  
As pointed-out by \citet{BenDavid-NIPS06}, a strategy to control the $\Hcal$-divergence is to find a representation of the examples where both the source and the target domain are as indistinguishable as possible. Under such a representation, a hypothesis with a low source risk will, according to Theorem~\ref{thm:RDT_bound}, perform well on the target data.  
In this paper, we present an algorithm that directly exploits this idea.

\section{Domain-Adversarial Neural Networks (DANN)}
\label{section:dann}

An original aspect of our approach is to explicitly implement the idea exhibited by Theorem~\ref{thm:RDT_bound} 
into a neural network classifier.
That is,  to learn a
model that can generalize well from one domain to another, we ensure that
the internal representation of the neural network contains no discriminative information about the origin of the input (source or target), while preserving a low risk on the source (labeled) examples.

In this section, we detail the proposed approach for incorporating a ``domain adaptation component'' to neural networks.
In Subsection~\ref{section:shallow_dann}, we start by developing the idea for the simplest possible case, \ie,  a single hidden layer, fully connected neural network. We then describe how to generalize the approach to arbitrary  (deep) network architectures. 

\subsection{Example Case with a Shallow Neural Network}
\label{section:shallow_dann}

Let us first consider a standard neural network (NN) architecture with a single hidden layer. For simplicity, we suppose that the input space is formed by $m$-dimensional real vectors. Thus, $X=\Rbb^m$. The hidden layer $G_f$ learns a function $G_f:X\to\Rbb^D$ that maps an example into a new $D$-dimensional representation\footnote{For brevity of notation, we will sometimes drop the dependence of $G_f$ on its parameters $(\WW,\bb)$ 
and shorten $G_f(\xb; \WW, \bb )$ to $G_f(\xb)$.}, and is parameterized by a matrix-vector pair $(\WW,\bb) \in \Rbb^{D\times m}\times \Rbb^D$\,: 
\begin{equation}
\label{eq:hh}
G_f(\xb; \WW, \bb )\, =\, \sigm\big(\WW\xx+\bb\big) \,,
\end{equation}
with \ $\sigm(\aaa) \, \eqdef\, \left[\tfrac{1}{1+\exp(-a_i)}\right]_{i=1}^{|\aaa|}$.\\

Similarly, the prediction layer $G_y$ learns a function $G_y:\Rbb^D \to [0,1]^L$ that is parameterized by a pair 
$(\VV,\cc)\in\Rbb^{L\times D}\times\Rbb^L$:
\begin{equation*}
G_y(G_f(\xb); \VV, \cc) 
\,=\, \softmax\big(\VV G_f(\xx)+\cc\big)\,,
\end{equation*}
with \ $\softmax(\aaa) \,\eqdef \,  \left[\frac{\exp(a_i)}{\sum_{j=1}^{|\aaa|}\exp(a_j)}\right]_{i=1}^{|\aaa|}$.\\

Here we have $L=|Y|$. By using the $\softmax$ function, each component of vector $G_y(G_f(\xx))$ denotes the conditional probability that the neural network assigns $\xx$ to the class in $Y$ represented by that component. 
Given a source example $(\xb_i,y_i)$, the natural classification loss to use is the negative log-probability of the correct label:
\begin{equation*}
\Lcal_y\big(G_y(G_f(\xx_i)),y_i\big) \ \eqdef \ \log\frac{1}{G_y(G_f(\xx))_{y_i}}\, .
\end{equation*}
Training the neural network then leads to the following optimization problem on the source domain:
\begin{equation} \label{eq:loss}
\min_{\WW, \bb, \VV, \cc} \,\Bigg[ \frac{1}{n} \sum_{i=1}^n \Lcal_y^i(\WW,\bb, \VV,\cc) + \lambda\cdot R(\WW,\bb)\Bigg]\,,
\end{equation}
where $\Lcal_y^i(\WW, \bb, \VV, \cc) = \Lcal_{y} \big( G_y(G_f(\xbs_i; \WW, \bb); \VV, \cc), y_i\big) $ is a shorthand notation for the prediction loss on the $i$-th example, and $R(\WW,\bb)$ is an optional regularizer that is weighted by hyper-parameter $\lambda$.
\medskip

The heart of our approach is to design a \emph{domain regularizer} directly derived from the $\Hcal$-divergence of Definition~\ref{def:Hdiv}. To this end, we view the output of the hidden layer $G_f(\cdot)$ (Equation~\ref{eq:hh}) as the internal representation of the neural network. 
Thus, we denote the source sample representations as 
$$S(G_f)\ \eqdef\  \big\{G_f(\xbs) \,\big|\, \xx\in S \big\}\,.$$
Similarly, given  an unlabeled sample from the target domain
we denote the corresponding representations 
$$T(G_f)\ \eqdef\  \big\{G_f(\xbt) \,\big|\, \xx\in T \big\}\,.$$
Based on Equation~\eqref{eq:Hdiv_empirique}, the empirical $\Hcal$-divergence of a symmetric hypothesis class $\Hcal$ between samples $S(G_f)$ and $T(G_f)$ is given by
\begin{equation} \label{eq:Hdiv_hh}
\hat{d}_\Hcal\big( S(G_f),T(G_f) \big) \,=\,
2\,\Bigg( 1 - \min_{\eta\in\Hcal} \bigg[
\frac{1}{n} \sum_{i=1}^n\! I\big[\eta(G_f(\xbs_i)){=} 0\big] 
+ \frac{1}{n'}\! \sum_{i=n+1}^{N}\! I\big[\eta(G_f(\xbt_i)){=}1\big]
\bigg] \Bigg).
\end{equation}
Let us consider $\Hcal$ as the class of hyperplanes in the representation space. Inspired by the Proxy $\Acal$-distance (see Section~\ref{section:PAD}), we suggest estimating the ``$\min$'' part of Equation~\eqref{eq:Hdiv_hh} by a \emph{domain classification layer} $G_d$ that learns a logistic regressor $G_d:\Rbb^D\to[0,1]$, parameterized by a vector-scalar pair $(\uu, z)\in \Rbb^D\times\Rbb$, that models the probability that a given input is from the source domain $\DSX$ or the target domain $\DTX$.
Thus,
\begin{equation} \label{eq:o}
G_d(G_f(\xb); \uu,z) \, \eqdef\, \sigm\big(\uu^\top G_f(\xb)+z\big)\,.
\end{equation}
Hence, the function $G_d(\cdot)$ is a \emph{domain regressor}.
We define its loss by
\begin{align*}  
\Lcal_d \big(G_d(G_f(\xb_i)), d_i\big)\ =\  
d_i \log\frac{1}{G_d(G_f(\xb_i))} + (1{-}d_i) \log\frac{1}{1{-}G_d(G_f(\xb_i))}\,,
\end{align*} 
where $d_i$ denotes the binary variable ({\em domain label}) for the $i$-th example, which indicates whether $\x_i$ come from the source distribution ($\x_i{\sim}\S$ if $d_i{=}0$) or from the target distribution ($\x_i{\sim}\T$ if $d_i{=}1$).

Recall that for the examples from the source distribution ($d_i{=}0$), the corresponding labels $y_i \in Y$ are known at training time. For the examples from the target domains, we do not know the labels at training time, and we want to predict such labels at test time.
This enables us to add a domain adaptation term to the objective of Equation~\eqref{eq:loss}, giving 
the following 
regularizer:
\begin{eqnarray}  \label{eq:opt-regul}
R(\WW,\bb) \ = \ 
\max_{\uu, z} \!
\left[
{-}\frac{1}{n} \sum_{i=1}^{n} \Lcal_{d}^i(\WW, \bb, \uu, z)  - \frac{1}{n'} \sum_{i=n+1}^{N} \Lcal_{d}^i(\WW, \bb, \uu, z  \big)
\right],
\end{eqnarray}
where
$ \Lcal_{d}^i(\WW, \bb, \uu, z) {=} \Lcal_{d} \big( G_d(G_f(\xbs_i; \WW, \bb); \uu, z), d_i)$. 
 This regularizer seeks to approximate the $\Hcal$-divergence 
of Equation~\eqref{eq:Hdiv_hh}, as 
 $2(1-R(\WW,\bb))$ is a surrogate for $\hat{d}_\Hcal\big( S(G_f),T(G_f) \big)$.
In line with Theorem~\ref{thm:RDT_bound}, the optimization problem given by Equations~\eqref{eq:loss} and~\eqref{eq:opt-regul} implements a trade-off between the minimization of the source risk $\RS(\cdot)$ and the divergence $\hat{d}_\Hcal( \cdot,\cdot)$. The hyper-parameter $\lambda$ is then used to tune the trade-off between these two quantities during the learning process.

For learning, we first note that we can rewrite the complete optimization objective of Equation~\eqref{eq:loss} as follows:
\begin{align} \label{eqn:global}
E (\WW, \VV, &\bb, \cc, \uu, z) \\
\nonumber
&= \ \frac{1}{n} \sum_{i=1}^n \Lcal_y^i(\WW,\bb, \VV,\cc)
- \lambda \,\Big(\frac{1}{n} \sum_{i=1}^{n} \Lcal_{d}^i(\WW, \bb, \uu, z)  + \frac{1}{n'} \sum_{\mathclap{i=n+1}}^{N} \Lcal_{d}^i(\WW, \bb, \uu, z )\Big),
\end{align}
where we are seeking the parameters $\hat{\WW}, \hat{\VV}, \hat{\bb}, \hat{\cc}, \hat{\uu}, \hat{z}$ that deliver a saddle point given by
\begin{eqnarray*}
(\hat{\WW}, \hat{\VV}, \hat{\bb}, \hat{\cc}) &=& \argmin_{\WW, \VV, \bb, \cc}\,E (\WW, \VV, \bb, \cc, \hat{\uu}, \hat{z})\,,\\
(\hat{\uu},\hat{z}) &=& \argmax_{\uu,z} \, E(\hat{\WW}, \hat{\VV}, \hat{\bb}, \hat{\cc}, \uu,z)\,.
\end{eqnarray*}
Thus, the optimization problem involves a minimization with respect to some parameters, as well as a maximization with respect to the others. 

\begin{algorithm}[t] \footnotesize
   \caption{Shallow DANN -- Stochastic training update}
   \label{alg:stoch-up}
\begin{multicols}{2}
\begin{algorithmic}[1]
   \STATE {\bfseries Input:} \\
   --- samples $S=\{(\xbs_i, \ys_i)\}_{i=1}^n$ and $T=\{\xbt_i\}_{i=1}^{n'}$,\\
   --- hidden layer size $D$, \\
   --- adaptation parameter $\lambda$,\\
   --- learning rate $\mu$,
   \STATE {\bfseries Output:} neural network $\{\WW, \VV, \bb, \cc\}$ 
   \vspace{2mm}
   \STATE $\WW, \VV \leftarrow {\rm random\_init}(\,D\,)$
   \STATE $\bb, \cc, \uu, d \leftarrow 0$
   \WHILE{stopping criterion is not met}
   \FOR{$i$ from 1 to $n$}
   \STATE \# {\tt Forward propagation}
   \STATE $G_f(\xbs_i) \leftarrow \sigm(\bb + \WW\xbs_i)$
   \STATE $G_y(G_f(\xbs_i)) \leftarrow \softmax(\cc + \VV G_f(\xbs_i))$
   \vspace{1.5mm}
   \STATE \# {\tt Backpropagation}
   \STATE $\Delta_{\cc} \leftarrow -(\ee(\ys_i)-G_y(G_f(\xbs_i)))$ 
   \STATE $\Delta_{\VV} \leftarrow \Delta_{\cc}~G_f(\xbs_i)^\top$ 
   \STATE $\Delta_{\bb} \leftarrow \left(\VV^{\top} \Delta_{\cc}\right) \odot G_f(\xbs_i) \odot (1-G_f(\xbs_i))$
   \STATE $\Delta_{\WW} \leftarrow \Delta_{\bb} \cdot({\xbs_i})^\top$
   \vspace{1.5mm}
   \STATE \# {\tt Domain adaptation regularizer...}
   \STATE \# {\tt ...from current domain}
   \STATE $G_d(G_f(\xbs_i)) \leftarrow \sigm(d + \uu^\top G_f(\xbs_i))$
   \STATE $\Delta_d \leftarrow \lambda (1-G_d(G_f(\xbs_i)))$ 
   \STATE $\Delta_{\uu} \leftarrow \lambda(1-G_d(G_f(\xbs_i))) G_f(\xbs_i)$
   \STATE ${\rm tmp} \leftarrow \lambda(1-G_d(G_f(\xbs_i)))$\\*
   ~~~~~~~~~~~~~~~${}\times\uu \odot G_f(\xbs_i) \odot (1-G_f(\xbs_i))$
   \STATE $\Delta_{\bb} \leftarrow \Delta_{\bb} + {\rm tmp}$ 
   \STATE $\Delta_{\WW} \leftarrow \Delta_{\WW} + {\rm tmp}\cdot({\xbs_i})^\top$
   \label{algoline:omit1}
      \vspace{1.5mm}
   \STATE \# {\tt ...from other domain}
   \STATE $j \leftarrow {\rm uniform\_integer}(1,\dots,n')$
   \STATE $G_f(\xbt_j) \leftarrow \sigm(\bb + \WW\xbt_j)$
   \STATE $G_d(G_f(\xbt_j)) \leftarrow \sigm(d + \uu^\top G_f(\xbt_j))$ 
   \STATE $\Delta_d \leftarrow \Delta_d - \lambda G_d(G_f(\xbt_j))$
   \STATE $\Delta_{\uu} \leftarrow \Delta_{\uu} - \lambda G_d(G_f(\xbt_j)) G_f(\xbt_j)$
   \STATE ${\rm tmp} \leftarrow -\lambda G_d(G_f(\xbt_j))$\\*
   ~~~~~~~~~~~~~~~${}\times\uu \odot G_f(\xbt_j) \odot (1-G_f(\xbt_j))$
   \STATE $\Delta_{\bb} \leftarrow \Delta_{\bb} + {\rm tmp}$ 
   \STATE $\Delta_{\WW} \leftarrow \Delta_{\WW} + {\rm tmp}\cdot(\xbt_j)^\top$
   \label{algoline:omit2}
      \vspace{1.5mm}
   \STATE \# {\tt Update neural network parameters}
   \STATE $\WW \leftarrow \WW - \mu \Delta_{\WW}$ 
   \STATE $\VV \leftarrow \VV - \mu \Delta_{\VV}$
   \STATE $\bb \leftarrow \bb - \mu \Delta_{\bb}$ 
   \STATE $\cc \leftarrow \cc - \mu \Delta_{\cc}$
      \vspace{1.5mm}
   \STATE \# {\tt Update domain classifier}
   \STATE $\uu \leftarrow \uu + \mu \Delta_{\uu}$ 
   \STATE $d \leftarrow d + \mu \Delta_{d}$
   \label{algoline:da}
      \ENDFOR
   \ENDWHILE
\end{algorithmic}
\end{multicols}
{\bf Note:} In this pseudo-code, $\ee(y)$ refers to a ``one-hot'' vector, consisting of all $0$s except for a $1$ at position~$y$, and $\odot$ is the element-wise product.
\end{algorithm}

We propose to tackle this problem with a simple stochastic gradient procedure, in which updates are made in the opposite direction of the gradient of Equation~\eqref{eqn:global} for the minimizing parameters, and in the direction of the gradient for the maximizing parameters. Stochastic estimates of the gradient are made, using a subset of the training samples to compute the averages. Algorithm~\ref{alg:stoch-up} provides the complete pseudo-code of this learning procedure.\footnote{We provide an implementation of \emph{Shallow DANN} algorithm at 
\url{http://graal.ift.ulaval.ca/dann/}}
In words, during training, the neural network (parameterized by $\WW, \bb, \VV, \cc$) and the domain regressor (parameterized by $\uu, z$) are competing
against each other, in an adversarial way, over the objective of Equation~\eqref{eqn:global}. 
For this reason, we refer to networks trained according to this objective
as Domain-Adversarial Neural Networks (DANN).
DANN will effectively attempt to learn a hidden layer $G_f(\cdot)$ that maps an example (either source or target) into a representation allowing the output layer $G_y(\cdot)$ to accurately classify source samples, but crippling the ability of the domain regressor $G_d(\cdot)$ to detect whether each example belongs to the source or target domains.

\subsection{Generalization to Arbitrary Architectures}
\label{section:deep_DANN}

For illustration purposes, we've so far focused on the case of a single hidden layer DANN. However, it is straightforward to generalize to other sophisticated architectures, which might be more appropriate for the data at hand. For example,  deep convolutional neural networks are well known for being state-of-the-art models for learning discriminative features of images~\citep{KrizhevskyA2012}.

Let us now use a more general notation for the different components of DANN. Namely, let $G_f(\cdot; \tf)$ be the $D$-dimensional neural network feature extractor, with parameters $\tf$. Also, let $G_y(\cdot; \ty)$ be the part of DANN that computes the network's label prediction output layer, with parameters $\ty$, while $G_d(\cdot;\td)$ now corresponds to the computation of the domain prediction output of the network, with parameters $\td$.
Note that for preserving the theoretical guarantees of Theorem~\ref{thm:RDT_bound}, the hypothesis class $\Hcal_d$ generated by the domain prediction component $G_d$ should include the hypothesis class $\Hcal_y$ generated by the label prediction component $G_y$. Thus, $\Hcal_y\subseteq\Hcal_d$.

We will note the prediction loss and the domain loss respectively by
\begin{eqnarray*}
\Lcal_y^i(\tf, \ty)&=& \Lcal_{y} \big( G_y(G_f(\xbs_i; \tf); \ty), y_i\big)\,,\\ 
\Lcal_{d}^i(\tf, \td)&=& \Lcal_{d} \big( G_d(G_f(\xbs_i; \tf); \td), d_i)\,.
\end{eqnarray*}
Training DANN then parallels the single layer case and consists in optimizing
\begin{eqnarray}
E (\tf,\ty,\td)
= \frac{1}{n}\sum_{i=1}^n \Lcal_y^i(\tf,\ty)
- \lambda \,\Big(\frac{1}{n} \sum_{i=1}^{n} \Lcal_{d}^i(\tf, \td)  + \frac{1}{n'} \sum_{\mathclap{i=n+1}}^{N} \Lcal_{d}^i(\tf, \td )\Big),
\label{eqn:global_gen}
\end{eqnarray}
by finding the saddle point $\htf,\hty,\htd$ such that
\begin{eqnarray}
(\htf,\hty) &=& \argmin_{\tf,\ty}\, E(\tf,\ty,\htd)\,,\label{eq:opt1}\\\label{eq:opt2}
\htd &=& \argmax_\td \, E(\htf,\hty, \td)\,.
\end{eqnarray}

As suggested previously, a saddle point defined by Equations~(\ref{eq:opt1}-\ref{eq:opt2}) can be found as a stationary point of the following gradient updates:
{\allowdisplaybreaks[4]
\begin{align}
\tf \quad &\longleftarrow \quad \tf \;-\; \mu \left(\frac{\partial \Lcal_y^i}{\partial \tf}-\lambda\frac{\partial \Lcal_d^i}{\partial \tf} \right), \label{eq:upd1}\\
\ty \quad &\longleftarrow \qquad \ty \;-\; \mu \frac{\partial \Lcal_y^i}{\partial \ty}\,,\label{eq:upd2}\\
\td \quad &\longleftarrow \qquad \td \;-\; \mu \lambda \frac{\partial \Lcal_d^i}{\partial \td}\,, \label{eq:upd3}
\end{align}
}where $\mu$ is the learning rate. We use stochastic estimates of these gradients, by sampling examples from the data set.

The updates of Equations~(\ref{eq:upd1}-\ref{eq:upd3}) are very similar to stochastic gradient descent (SGD) updates for a feed-forward deep model that comprises feature extractor fed into the label predictor and into the domain classifier (with loss weighted by $\lambda$). The only difference is that in \eq{upd1}, the gradients from the class and domain predictors are subtracted, instead of being summed (the difference is important, as otherwise SGD would try to make features dissimilar across domains in order to minimize the domain classification loss). Since SGD---and its many variants, such as ADAGRAD~\citep{Duchi:EECS-2010-24} or ADADELTA~\citep{Zeiler2012}---is the main learning algorithm implemented in most libraries for deep learning, it would be convenient to frame an implementation of our stochastic saddle point procedure as SGD.

Fortunately, such a reduction can be accomplished by introducing a special \emph{gradient reversal layer} (GRL), defined as follows. The gradient reversal layer has no parameters associated with it.
During the forward propagation, the GRL acts as an identity transformation. During the backpropagation however, the GRL takes the gradient from the subsequent level and changes its sign, \ie,  multiplies it by $-1$, before passing it to the preceding layer. Implementing such a layer using existing object-oriented packages for deep learning is simple, requiring only to define procedures for the forward propagation (identity transformation), and backpropagation (multiplying by $-1$). The layer requires no parameter update. 

\begin{figure*}[t]
 \centering
 \includegraphics[width=0.85\textwidth]{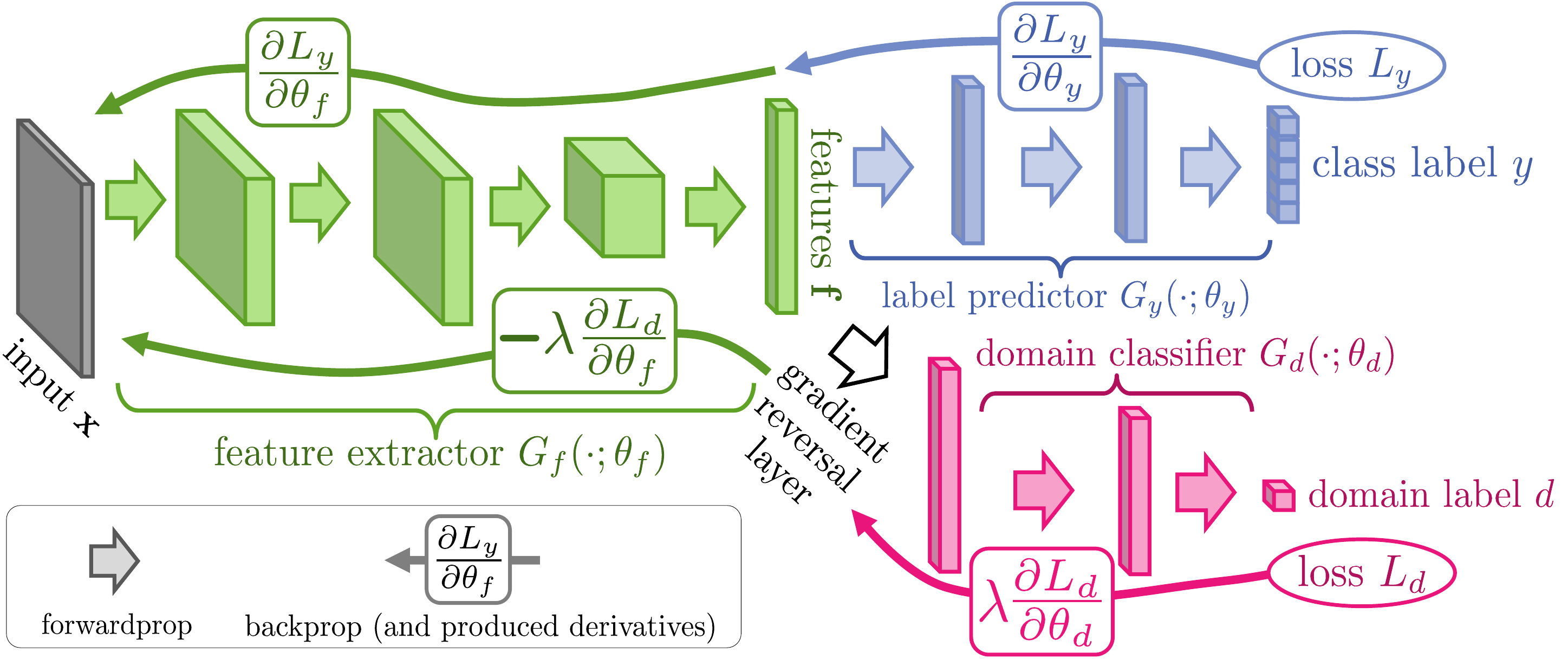}
 \caption{The {\bf proposed architecture} includes a deep {\em feature extractor} (green) and a deep {\em label predictor} (blue), which together form a standard feed-forward architecture. Unsupervised domain adaptation is achieved by adding a {\em domain classifier} (red) connected to the feature extractor via a {\em gradient reversal layer} that multiplies the gradient by a certain negative constant during the backpropagation-based training. Otherwise, the training proceeds standardly and minimizes the label prediction loss (for source examples) and the domain classification loss (for all samples). Gradient reversal ensures that the feature distributions over the two domains are made similar (as indistinguishable as possible for the domain classifier), thus resulting in the domain-invariant features.}
 \label{fig:arch}
 \end{figure*}

The GRL as defined above is inserted between the feature extractor $G_f$ and the domain classifier $G_d$, resulting in the architecture depicted in \fig{arch}. As the backpropagation process passes through the GRL, the partial derivatives of the loss that is downstream the GRL (\ie, $\Lcal_d$) w.r.t.\ the layer parameters that are upstream the GRL (\ie, $\tf$) get multiplied by $-1$, \ie, $\frac{\partial \Lcal_d}{\partial \tf}$ is effectively replaced with $-\frac{\partial \Lcal_d}{\partial \tf}$. Therefore, running SGD in the resulting model implements the updates of Equations~(\ref{eq:upd1}-\ref{eq:upd3}) and converges to a saddle point of Equation~\eqref{eqn:global_gen}.

Mathematically, we can formally treat the gradient reversal layer as a ``pseudo-function'' ${\cal R}(\x)$ defined by two (incompatible) equations describing its forward and backpropagation behaviour:
\begin{align}
{\cal R}(\x) = \x\,,\\
\frac{d{\cal R}}{d\x} = -\mathbf{I}\,,
\end{align}
where $\mathbf{I}$ is an identity matrix.
We can then define the objective ``pseudo-function'' of $(\tf,\ty,\td)$ that is being optimized by the stochastic gradient descent within our method:
\begin{eqnarray}
&&\tilde E (\tf,\ty,\td) = \! \frac{1}{n}\sum_{i=1}^n \Lcal_y\left( \strut G_y(G_f(\x_i;\tf);\ty), y_i\right)\label{eq:pseudoobj}\\
&&~~~- \lambda \,\Big(\frac{1}{n} \sum_{i=1}^{n} \Lcal_d \left( \strut G_d({\cal R}(G_f(\x_i;\tf));\td), d_i\right) + \frac{1}{n'} \sum_{\mathclap{i=n+1}}^{N} \Lcal_d \left( \strut G_d({\cal R}(G_f(\x_i;\tf));\td), d_i\right) \Big)\,.\nonumber
\end{eqnarray}

Running updates (\ref{eq:upd1}-\ref{eq:upd3}) can then be implemented as doing SGD for \eq{pseudoobj} and leads to the emergence of features that are domain-invariant and discriminative at the same time. After the learning, the label predictor $G_y(G_f(\x;\tf);\ty)$ can be used to predict labels for samples from the target domain (as well as from the source domain).
Note that we release the source code for
the Gradient Reversal layer along with the usage examples
as an extension to \texttt{Caffe} \citep{Jia14}.\footnote{\url{http://sites.skoltech.ru/compvision/projects/grl/}}

\section{Experiments}
In this section, we present a variety of empirical results for both \emph{shallow} domain adversarial neural networks (Subsection~\ref{section:experiments_shallow}) and \emph{deep} ones (Subsections~\ref{sect:experiments} and~\ref{section:deep_image}). 
 
\subsection{Experiments with Shallow Neural Networks}
\label{section:experiments_shallow}

In this first experiment section, we evaluate the behavior of the simple version of DANN described by Subsection~\ref{section:shallow_dann}. Note that the results reported in the present subsection are obtained using Algorithm~\ref{alg:stoch-up}.
Thus, the stochastic gradient descent approach here consists of sampling a pair of source and target examples and performing a gradient step update of all parameters of DANN. Crucially, while the update of the regular parameters follows as usual the opposite direction of the gradient, for the adversarial parameters the step must follow the gradient's direction (since we maximize with respect to them, instead of minimizing).

\subsubsection{Experiments on a Toy Problem}
\label{sec:toy_problem}

\begin{figure*}[t]
\centering
\subfloat[Standard NN. For the ``domain classification'', we use a \emph{non adversarial} domain regressor on the hidden neurons learned by the Standard NN. (This is equivalent to run Algorithm~\ref{alg:stoch-up}, without Lines~\ref{algoline:omit1} and~\ref{algoline:omit2})]
{\footnotesize\bf\sc
\begin{minipage}{.245\textwidth}\centering
Label classification\\[1mm]
\includegraphics[width=1\textwidth]{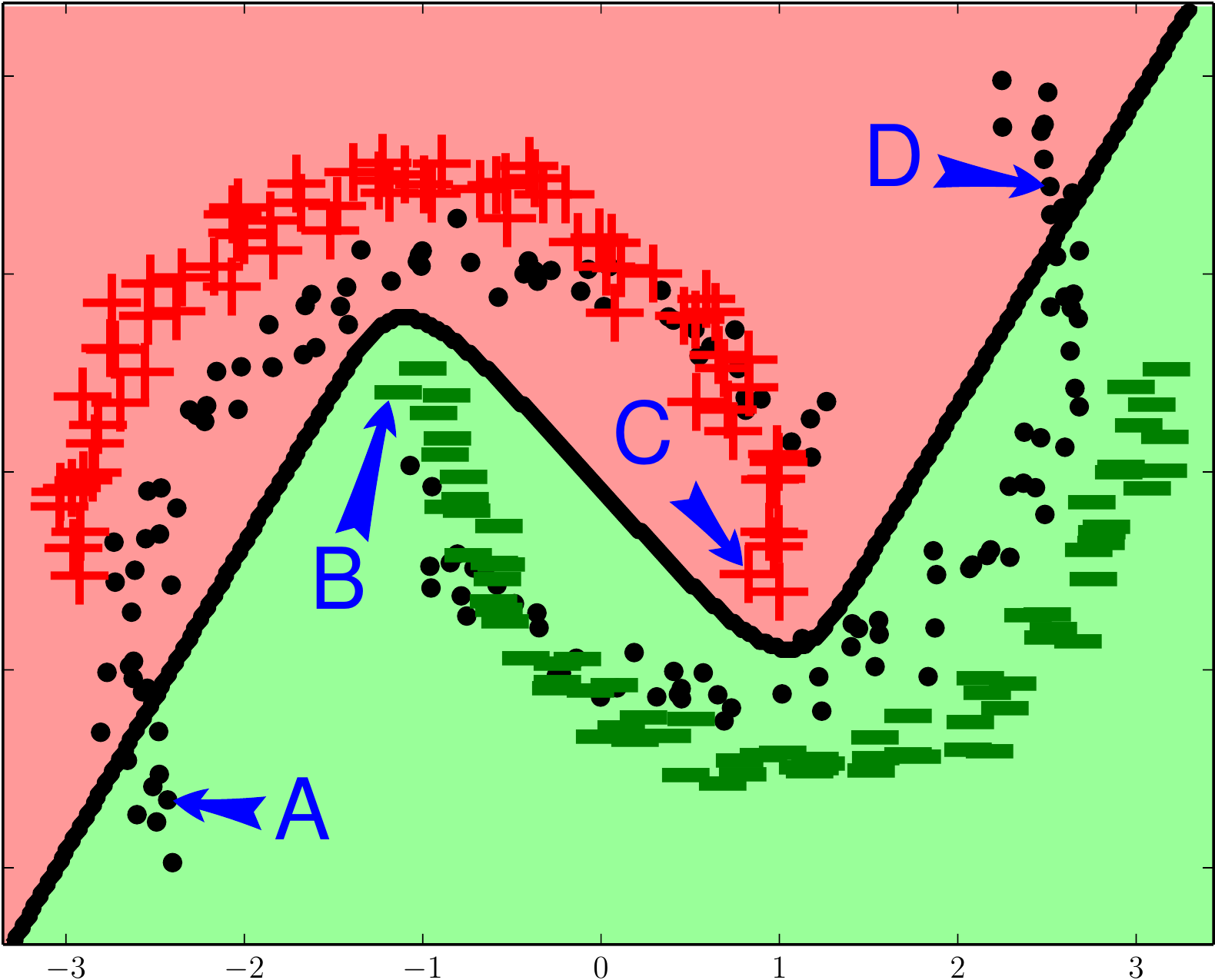}
\end{minipage}
\begin{minipage}{.245\textwidth}\centering
Representation PCA\\[1mm]
\includegraphics[width=1\textwidth]{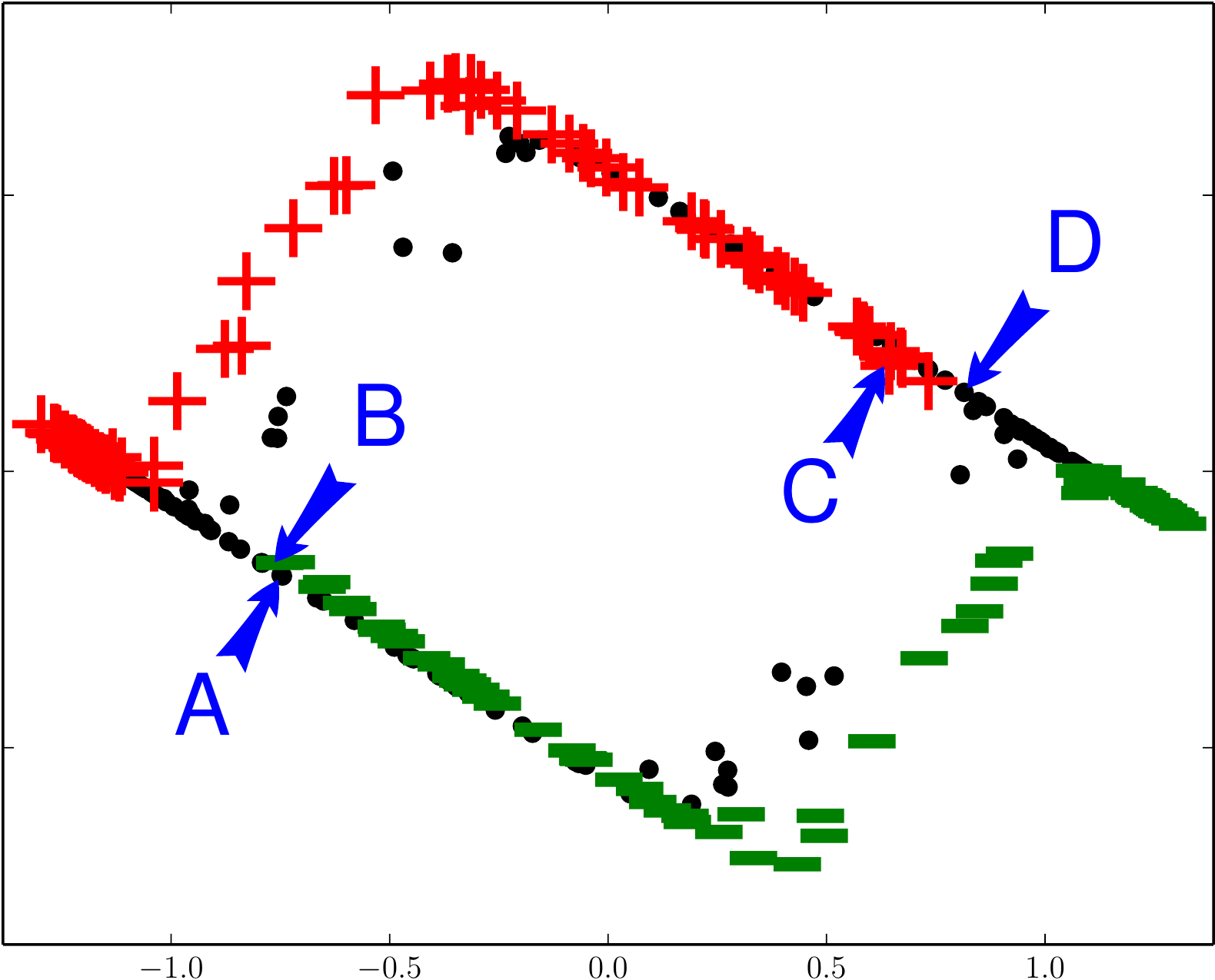}
\end{minipage}
\begin{minipage}{.245\textwidth}\centering
Domain classification\\[1mm]
\includegraphics[width=1\textwidth]{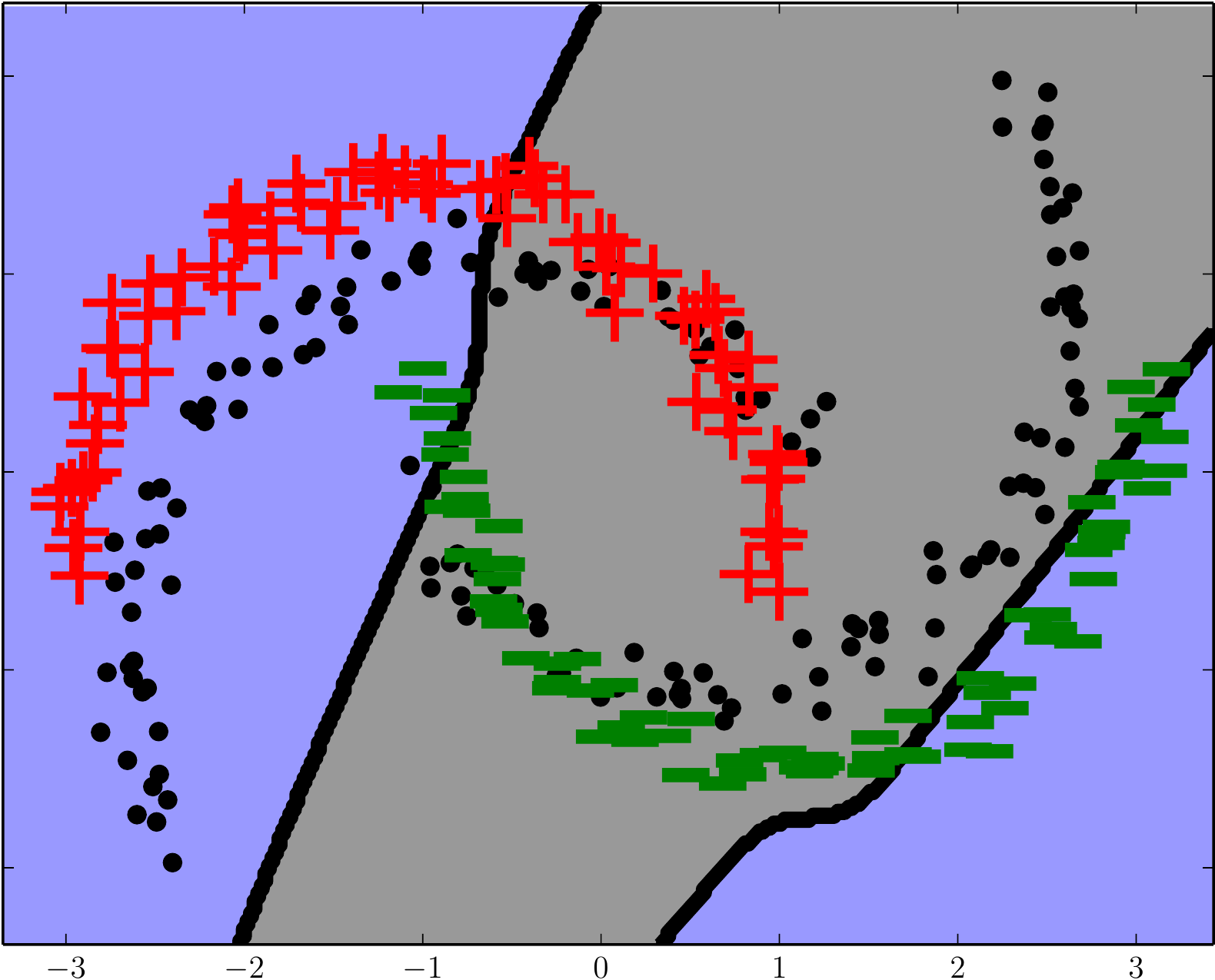}
\end{minipage}
\begin{minipage}{.245\textwidth}\centering
Hidden neurons\\[1mm]
\includegraphics[width=1\textwidth]{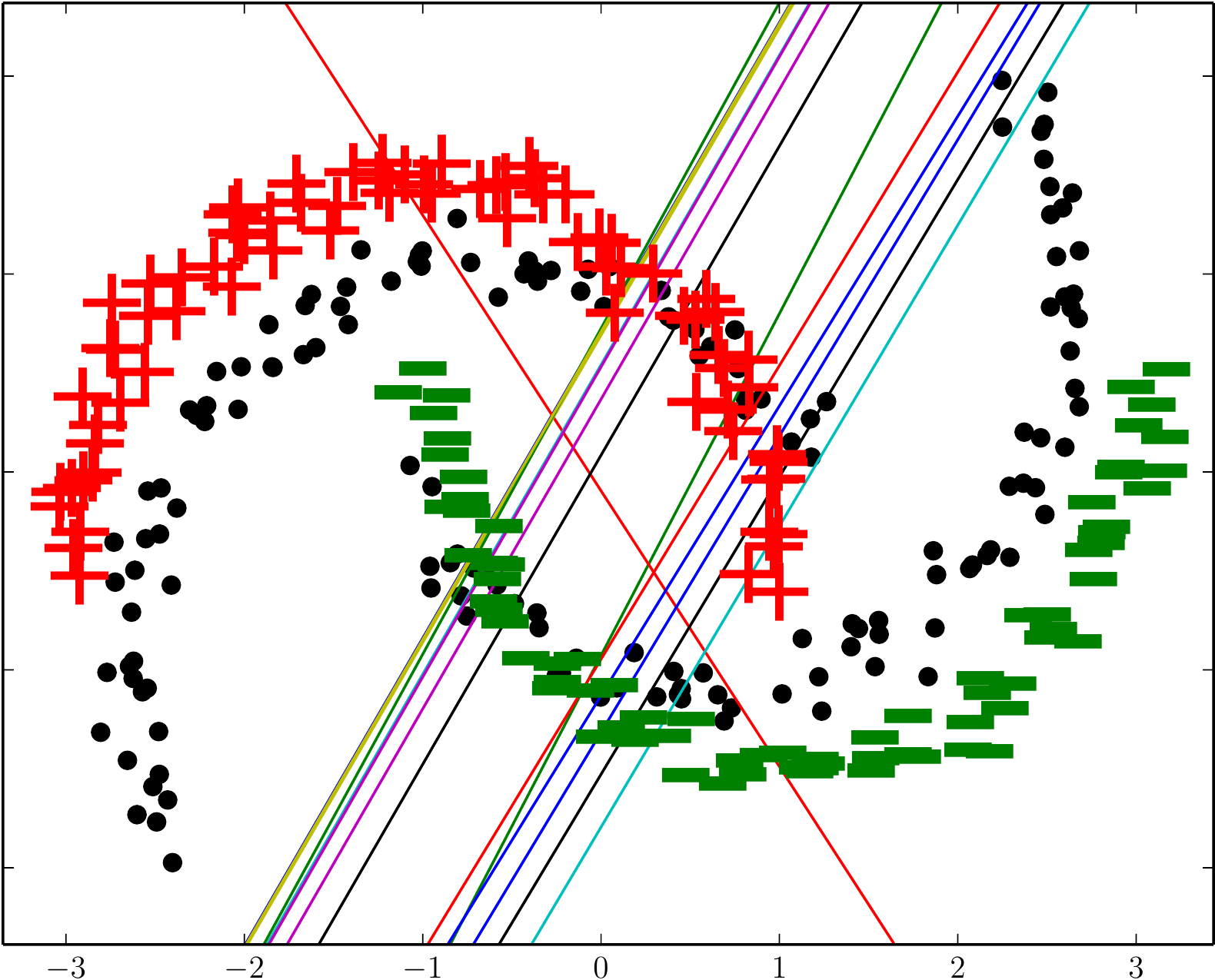}
\end{minipage} 
\label{fig:2moons_NN}
 }\\
 \subfloat[DANN (Algorithm~\ref{alg:stoch-up})]
 {\footnotesize\bf\sc
 \begin{minipage}{.245\textwidth}\centering
 \includegraphics[width=1\textwidth]{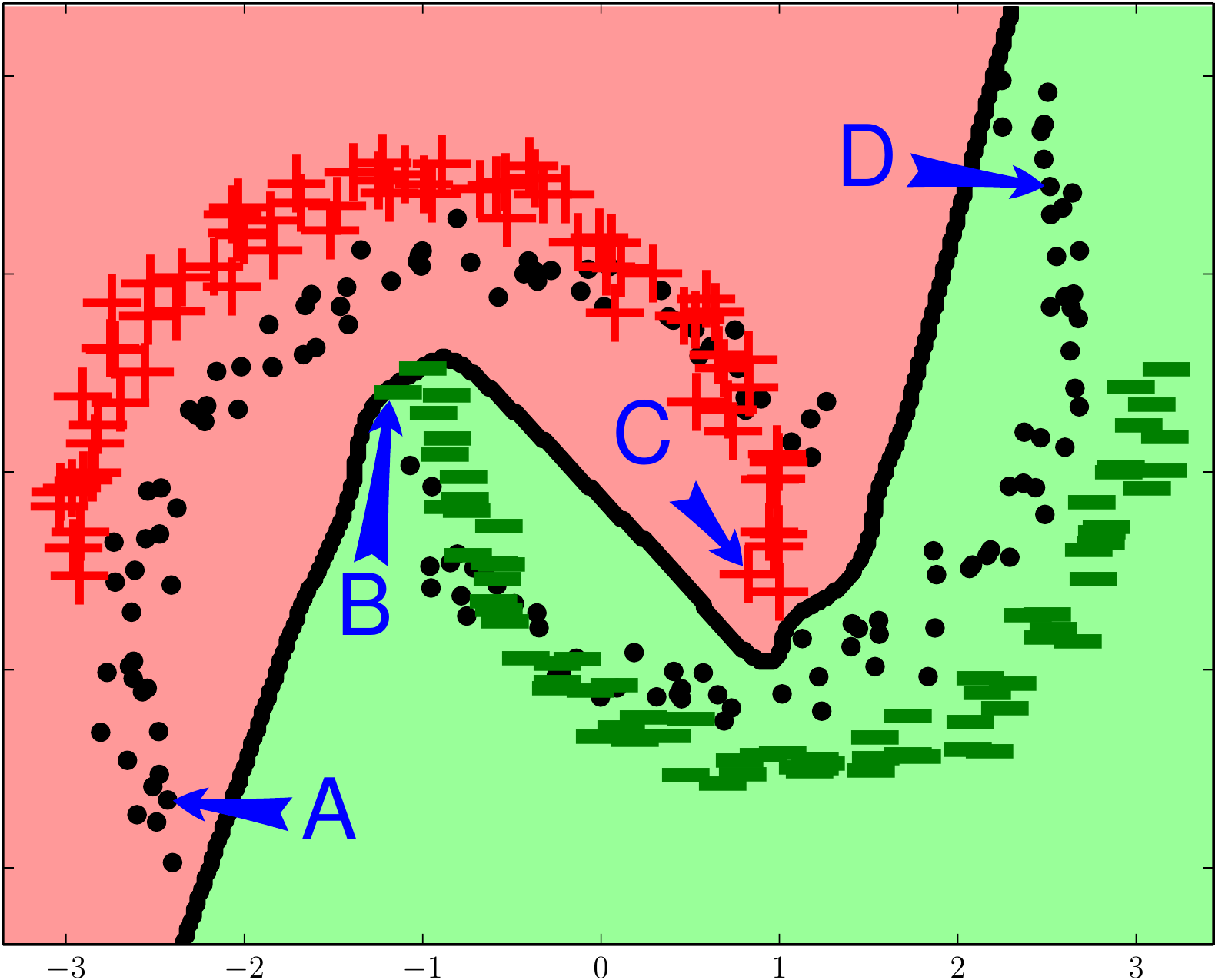}
 \end{minipage}
 \begin{minipage}{.245\textwidth}\centering
 \includegraphics[width=1\textwidth]{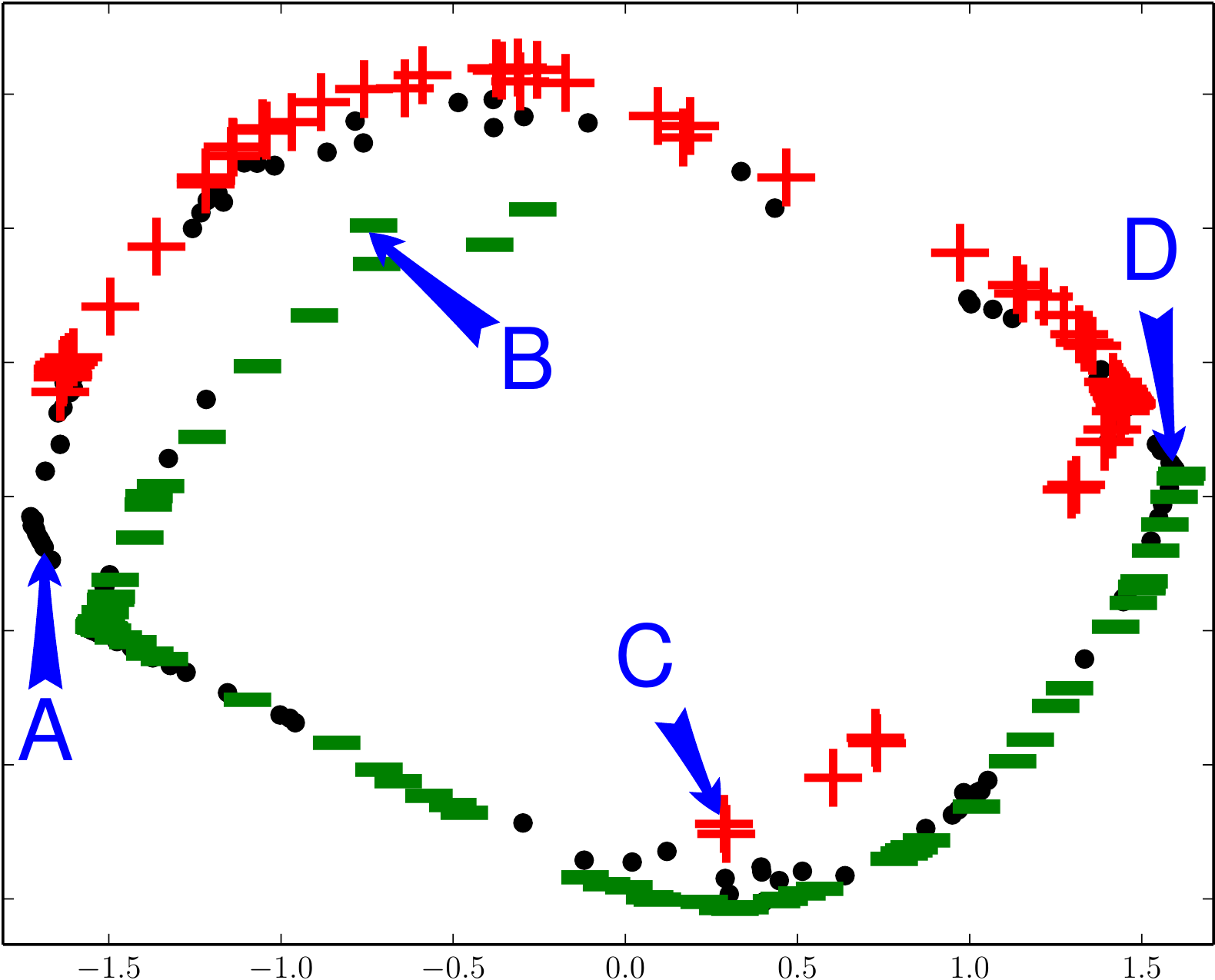}
 \end{minipage}
 \begin{minipage}{.245\textwidth}\centering
 \includegraphics[width=1\textwidth]{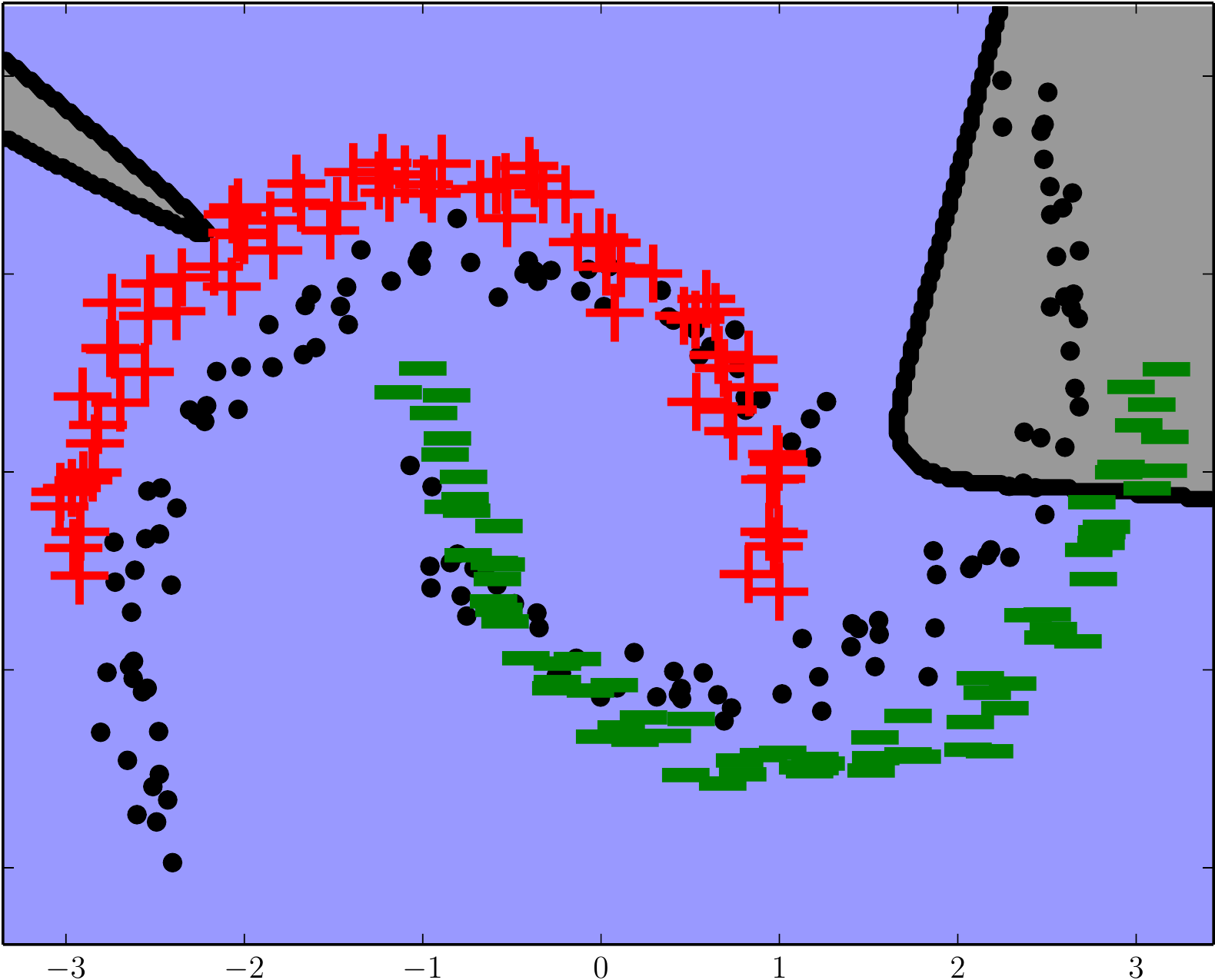}
 \end{minipage}
 \begin{minipage}{.245\textwidth}\centering
 \includegraphics[width=1\textwidth]{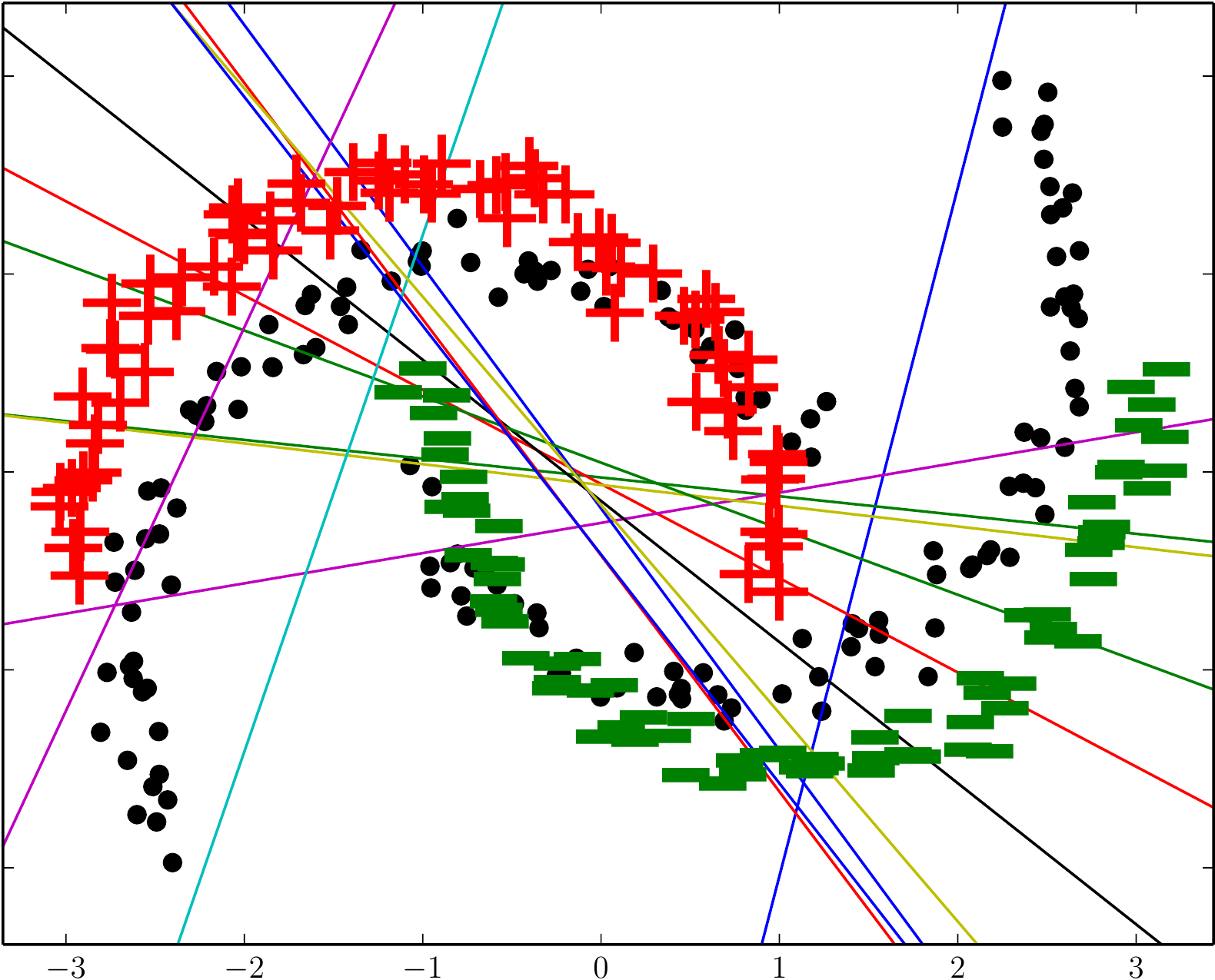}
 \end{minipage}
 \label{fig:2moons_DANN}
  }
\caption{The \emph{inter-twinning moons} toy problem. 
Examples from the source sample are represented as a~\redplus (label $1$) and a~\greenminus (label~$0$), while examples from the unlabeled target sample are represented as black dots. See text for the figure discussion.
\label{fig:2moons} }
\end{figure*}

As a first experiment, we study the behavior of the proposed algorithm on a variant of the \emph{inter-twinning moons} 2D problem, where the target distribution is a rotation of the source one. As the source sample $S$, we generate a lower moon and an upper moon labeled~$0$ and~$1$ respectively, each of which containing $150$ examples.  The target sample $T$ is obtained by the following procedure: (1) we generate a sample $S'$ the same way $S$ has been generated; (2) we rotate each example by $35\degree$; and (3) we remove all the labels. Thus, $T$ contains $300$ unlabeled examples. We have represented those examples in Figure~\ref{fig:2moons}.

We study the adaptation capability of DANN by comparing it to the standard neural network (NN).
In these toy experiments, both algorithms share the same network architecture, with a hidden layer size of $15$ neurons. We train the NN using the same procedure as the DANN. That is, we keep updating the domain regressor component using target sample $T$ (with a hyper-parameter $\lambda=6$; the same value is used for DANN), but we disable the \emph{adversarial} back-propagation into the hidden layer. To do so, we execute Algorithm~\ref{alg:stoch-up} by omitting the lines numbered~\ref{algoline:omit1} and~\ref{algoline:omit2}.
This allows recovering the NN learning algorithm---based on the source risk minimization of Equation~\eqref{eq:loss} without any regularizer---and simultaneously train the domain regressor of Equation~\eqref{eq:o} to discriminate between source and target domains. With this toy experience, we will first illustrate how DANN adapts its decision boundary when compared to NN.
Moreover, we will also illustrate how the representation given by the hidden layer is less adapted to the source domain task with DANN than with NN (this is why we need a domain regressor in the NN experiment). 
We recall that this is the founding idea behind our proposed algorithm.
The analysis of the experiment appears in Figure~\ref{fig:2moons}, where upper graphs relate to standard NN, and lower graphs relate to DANN. By looking at the lower and upper graphs pairwise, we compare NN and DANN from four different perspectives, described in details below.
\smallskip

The column ``\textsc{Label Classification}'' of Figure~\ref{fig:2moons} shows the decision boundaries of DANN and NN on the problem of predicting the labels of both source and the target examples. 
As expected, NN  accurately classifies the two classes of the source sample $S$, but is  \emph{not fully adapted} to the target sample $T$.
On the contrary, the decision boundary of DANN perfectly classifies examples from both source and target samples. In the studied task, DANN clearly adapts to the target distribution.
\smallskip

The column ``\textsc{Representation PCA}'' 
studies how the domain adaptation regularizer affects the representation $G_f(\cdot)$ provided by the network hidden layer. The graphs are obtained by applying a Principal component analysis (PCA) on the set of all representation of source and target data points, \ie, $S(G_f)\cup T(G_f)$. Thus, given the trained network (NN or DANN), every point from $S$ and~$T$ is mapped into a $15$-dimensional feature space through the hidden layer, and projected back into a two-dimensional plane by the PCA transformation. 
In the DANN-PCA representation, we observe that target points are homogeneously spread out among source points; In the NN-PCA representation, a number of target points belong to clusters containing no source points. Hence, labeling the target points seems an easier task given the DANN-PCA representation.\\
To push the analysis further, the PCA graphs tag four crucial data points by the letters A, B, C and D, that correspond to the moon extremities in the original space (note that the original point locations are tagged in the first column graphs). We observe that points A and B are very close to each other in the NN-PCA representation, while they clearly belong to different classes. The same happens to points C and D. Conversely, these four points are at the opposite four corners in the DANN-PCA representation. Note also that the target point A (resp. D)---that is difficult to classify in the original space---is located in the~\redplus cluster (resp.~\greenminus cluster) in the DANN-PCA representation.
Therefore, the representation promoted by DANN is better suited to the adaptation problem.
\smallskip

The column ``\textsc{Domain Classification}'' 
shows the decision boundary on the domain classification problem, which is given by  the  domain regressor $G_d$ of Equation~\eqref{eq:o}. More precisely, an example $\xb$ is classified as a source example when $G_d(G_f(\xb))\geq 0.5$, and is classified as a domain example otherwise. Remember that, during the learning process of DANN, the $G_d$ regressor struggles to discriminate between source and target domains, while the hidden representation $G_f(\cdot)$ is \emph{adversarially} updated to prevent it to succeed. As explained above, we trained a domain regressor during the learning process of NN, but without allowing it to influence the learned representation $G_f(\cdot)$.\\
On one hand, the DANN domain regressor clearly fails to generalize source and target distribution topologies. On the other hand, the NN domain regressor shows a better (although imperfect) generalization capability. Inter alia, it seems to roughly capture the rotation angle of the target distribution.  This again corroborates that the DANN representation does not allow discriminating between domains.
\smallskip

The column ``\textsc{Hidden Neurons}'' shows the configuration of hidden layer neurons 
(by Equation~\ref{eq:hh}, we have that each neuron is indeed a linear regressor). In other words, each of the fifteen plot line corresponds to the coordinates $\xb\in\Rbb^2$ for which the $i$-th component of $G_f(\xb)$  equals~$\frac{1}{2}$, for $i\in\{1,\ldots,15\}$. 
We observe that the standard NN neurons are grouped in three clusters, each one allowing to generate a straight line of the \emph{zigzag} decision boundary for the label classification problem. However, most of these neurons are also able to (roughly) capture the rotation angle of the domain classification problem. Hence, we observe that the adaptation regularizer of DANN prevents these kinds of neurons to be produced. It is indeed striking to see that the two predominant patterns in the NN neurons (\ie, the two parallel lines crossing the plane from lower left to upper right) are vanishing in the DANN neurons.

\subsubsection{Unsupervised Hyper-Parameter Selection}
\label{section:shallow_model_selection}

To perform unsupervised domain adaption, one should provide ways to set hyper-parameters (such as the domain regularization parameter $\lambda$, the learning rate, the network architecture for our method) in an unsupervised way, \ie, without referring to labeled data in the target domain. 
In the following experiments of Sections~\ref{section:expe_original_data} and~\ref{section:autoencoders},
we select the hyper-parameters of each algorithm by using a variant of \emph{reverse cross-validation} approach proposed by~\citet{zhong2010cross}, that we call \emph{reverse validation}.

To evaluate the \emph{reverse validation risk} associated to a tuple of hyper-parameters, we proceed as follows.
Given the labeled source sample $S$ and the  unlabeled target sample $T$, we split each set into training sets ($S'$ and $T'$ respectively, containing $90\%$ of the original examples) and the validation sets ($S_V$ and $T_V$ respectively). 
We use the labeled  set $S'$ and the unlabeled target set $T'$ to learn a classifier $\eta$. Then, using the same algorithm, we learn a \emph{reverse} classifier $\eta_r$ using the \emph{self-labeled} set $\{(\xx, \eta(\xx))\}_{\xx\in T'}$ and the unlabeled part of $S'$ as target sample. Finally, the reverse classifier $\eta_r$ is evaluated on the validation set $S_V$ of source sample. 
We then say that the classifier $\eta$ has a \emph{reverse validation risk} of
$R_{S_V}(\eta_r)$. The process is repeated with multiple values of hyper-parameters and
the selected parameters are those corresponding to the classifier
with the lowest reverse validation risk. 

Note that when we train neural network architectures, the validation set $S_V$ is also used as an early stopping criterion during the learning of $\eta$, and \emph{self-labeled} validation set $\{(\xx, \eta(\xx))\}_{\x\in T_V}$ is used as an early stopping criterion during the learning of $\eta_r$. 
We also observed better accuracies when we initialized the learning of the reverse classifier $\eta_r$ with the configuration learned by the network $\eta$.

\subsubsection{Experiments on Sentiment Analysis Data Sets} 
\label{section:expe_original_data}

\begin{table*}[t]
\centering
\small \subfloat[Classification accuracy on the Amazon reviews data set]{\small\sc\renewcommand{\arraystretch}{1.2}\rowcolors{5}{}{black!10}
\begin{tabular}{llcccccc}
\toprule
\multicolumn{2}{c}{  } & \multicolumn{3}{c}{\textbf{Original data}}  & \multicolumn{3}{c}{\textbf{mSDA representation}} \\
 \cmidrule(l r){3-5} \cmidrule(l r){6-8}
 Source  & Target  & DANN & NN  & SVM & DANN &  NN &  SVM  \\
\cmidrule(l r){1-2} \cmidrule(l r){3-5} \cmidrule(l r){6-8}
 
books & dvd &.784	 &.790	     &\textbf{.799} &.829	 &.824&\textbf{.830} 	 \\ 
books & electronics    &.733 &.747 &	\textbf{.748} 			 & \textbf{.804} 		 &		.770	 &		.766 \\ 
books & kitchen    & \textbf{.779} 		 &	.778		 &		.769		 & \textbf{.843} 		 &		.842	 &		.821 \\ 
dvd & books      & .723	 &	.720		 &	\textbf{.743} 			 &	.825	 &		.823	 &	\textbf{.826} 	 \\ 
dvd & electronics   & \textbf{.754} 		 &	.732		 &		.748		 & \textbf{.809} 		 &		.768	 &		.739 \\ 
dvd & kitchen    & \textbf{.783} 		 &	.778		 &		.746		 &	.849	 &	\textbf{.853} 		 &		.842 \\ 
electronics & books   &	\textbf{.713} 	 &	.709		 &	    .705		 & \textbf{.774} 		 &		.770	 &		.762 \\ 
electronics & dvd     &	\textbf{.738} 	 &	.733		 &		.726		 &	\textbf{.781} 	 &		.759	 &		.770 \\ 
electronics & kitchen    & \textbf{.854} 		 &\textbf{.854} 			 &		.847		 &	.881	 &	\textbf{.863} 		 &		.847 \\ 
kitchen & books  &	\textbf{.709} 	 &	.708		 &		.707		 &	.718	 &		.721	 &	\textbf{.769} 	\\ 
kitchen & dvd   & \textbf{.740} 		 &	.739		 &		.736		 & \textbf{.789} 		 &	\textbf{.789} 		 & .788 \\ 
kitchen & electronics   &\textbf{.843}	 &	.841		 &	.842	 &	.856	 &		.850	 &	\textbf{.861} 	 \\

\bottomrule
\end{tabular}

\label{table:risks_a}}

\subfloat[Pairwise Poisson binomial test]{\small\sc\begin{tabular}{lccc}
\toprule
 \multicolumn{4}{c}{\textbf{Original data}}  \\
 \midrule
 & DANN & NN  & SVM   \\[1mm]

DANN        & .50 &  \textbf{.87} & \textbf{.83}  \\[1mm]

NN           & .13 & .50 & .63   \\[1mm]

SVM            & .17  & .37  & .50  \\

\bottomrule
\end{tabular}
\qquad
\begin{tabular}{lccc}
\toprule
 \multicolumn{4}{c}{\textbf{mSDA representations}}  \\
 \midrule
 & DANN & NN  & SVM   \\[1mm]
 
DANN      &    .50 &  \textbf{.92}  & \textbf{.88}  \\[1mm]

NN       &   .08   & .50 & .62 \\ [1mm]

SVM            & .12 & .38 &  .50 \\
\bottomrule
\end{tabular}
\label{table:risks_b}}
\caption{Classification accuracy on the Amazon reviews data set, and Pairwise Poisson binomial test. }  \label{table:risks}
\end{table*}

We now compare the performance of our proposed DANN algorithm to a standard neural network with one hidden layer (NN) described by Equation~\eqref{eq:loss}, and a Support Vector Machine (SVM) with a linear kernel.
We compare the algorithms on the {\it Amazon reviews} data set, as pre-processed by~\citet{Chen12}. 
This data set includes four domains, each one composed of reviews of a specific kind of product (books, dvd disks, electronics, and kitchen appliances). Reviews are encoded in $5\,000$ dimensional feature vectors of unigrams and bigrams, and labels are binary: ``$0$'' if the product is ranked up to $3$ stars, and ``$1$'' if the product is ranked $4$ or $5$ stars.

We perform twelve domain adaptation tasks. 
All learning algorithms are given  $2\,000$ labeled source examples and $2\,000$ unlabeled target examples. Then, we evaluate them on separate target test sets (between $3\,000$ and $6\,000$ examples).
Note that NN and SVM do not use the unlabeled target sample for learning. 

Here are more details about the procedure used for each learning algorithms leading to the empirical results of Table~\ref{table:risks}.
\begin{itemize}
\item For the {\sc DANN} algorithm, the adaptation parameter $\lambda$ is chosen among 9 values between $10^{-2}$ and $1$ on a logarithmic scale. The hidden layer size $l$ is either $50$ or~$100$. Finally, the learning rate $\mu$ is fixed at $10^{-3}$.
\item For the {\sc NN} algorithm, we use exactly the same hyper-parameters grid and training procedure as DANN above, except that we do not need an adaptation parameter. Note that one can train NN by using the DANN implementation (Algorithm~\ref{alg:stoch-up}) with $\lambda=0$.
\item For the {\sc SVM} algorithm, the hyper-parameter $C$ is chosen among 10 values between $10^{-5}$ and~$1$ on a logarithmic scale. This range of values is the same as used by~\citet{Chen12} in their experiments.
\end{itemize}
As presented at Section~\ref{section:shallow_model_selection}, we used 
\emph{reverse cross validation} selecting the hyper-parameters for all three learning algorithms, with \emph{early stopping} as the stopping criterion for {\sc DANN} and {\sc NN}.

The ``Original data'' part of Table~\ref{table:risks_a} shows
the target test accuracy of all algorithms, and Table~\ref{table:risks_b}
reports the probability that one algorithm is significantly better than the others according to the Poisson binomial test~\citep{lacoste-2012}.
We note that DANN has a significantly better performance than NN and SVM, with respective probabilities \textbf{0.87} and~\textbf{0.83}.  As the only difference between DANN and NN is the domain adaptation 
regularizer, we conclude that our approach successfully helps to find a representation suitable for the target domain.

\subsubsection{Combining DANN with Denoising Autoencoders}
\label{section:autoencoders}
 
We now investigate on whether the DANN algorithm can improve on the representation learned by the state-of-the-art \emph{Marginalized Stacked Denoising Autoencoders} (mSDA) proposed by~\citet{Chen12}.
In brief, mSDA is an unsupervised algorithm that learns a new  robust feature representation of the training samples. 
It takes the unlabeled parts of both source and target samples 
to learn a feature map from input space $\Xcal$ to a new representation space. 
As a \emph{denoising autoencoders} algorithm, it finds a feature representation from which one can (approximately) reconstruct
the original features of an example from its noisy counterpart.
\citet{Chen12} showed that using mSDA with a linear SVM classifier reaches state-of-the-art performance on the \emph{Amazon reviews} data sets. As an alternative to the SVM, we propose to apply our Shallow DANN algorithm on the same representations generated by mSDA (using representations of both source and target samples). Note that, even if mSDA and DANN are two representation learning approaches, they optimize different objectives, which can be complementary. 

We perform this experiment on the same \emph{Amazon reviews} data set described in the previous subsection. For each source-target domain pair, we generate the mSDA representations using a corruption probability of~$50\%$ and a number of layers of~$5$.  We then execute the three learning algorithms (DANN, NN, and SVM) on these representations. 
More precisely, following the experimental procedure of \citet{Chen12}, we use the concatenation of the output of the~$5$ layers and the original input as the new representation. Thus, each example is now encoded in a vector of $30\,000$ dimensions.
Note that we use the same grid search as in the previous Subsection~\ref{section:expe_original_data}, but use a learning rate $\mu$ of $10^{-4}$ for both DANN and the NN.
The results of ``mSDA representation'' columns in Table~\ref{table:risks_a}  confirm that combining mSDA and DANN is a sound approach. Indeed, the Poisson binomial test shows that DANN has a better performance than the NN and the SVM, with probabilities \textbf{0.92} and \textbf{0.88} respectively, as reported in Table~\ref{table:risks_b}.
We note however that the standard NN and the SVM find the best solution on respectively the second and the fourth tasks. This suggests that DANN and mSDA adaptation strategies are not fully complementary. 

\subsubsection{Proxy Distance} 
\label{section:PAD_experiments}

The theoretical foundation of the DANN algorithm is the domain adaptation theory of \citet{BenDavid-NIPS06,BenDavid-MLJ2010}. We claimed that DANN finds a representation in which the source and the target example are hardly distinguishable. Our toy experiment of Section~\ref{sec:toy_problem} already points out some evidence for that and here we provide analysis on real data. To do so, we compare the Proxy $\Acal$-distance (PAD) on various representations of the \emph{Amazon Reviews} data set; these representations are obtained by running either NN, DANN, mSDA, or mSDA and DANN combined. Recall that PAD, as described in Section~\ref{section:PAD}, is a metric estimating the similarity of the source and the target representations. 
More precisely, to obtain a PAD value, we use the following procedure: (1) we construct the data set $U$ of Equation~\eqref{eq:U} using both source and target representations of the training samples; (2) we randomly split $U$ in two subsets of equal size; (3) we train linear SVMs on the first subset of~$U$ using a large range of $C$ values; (4) we compute the error of all obtained classifiers on the second subset of~$U$; and (5) we use the lowest error to compute the PAD value of Equation~\eqref{eq:PAD}.

\begin{figure*}[t]
\centering
\subfloat[DANN on \emph{Original data}.]{
\includegraphics[width=.31\textwidth]{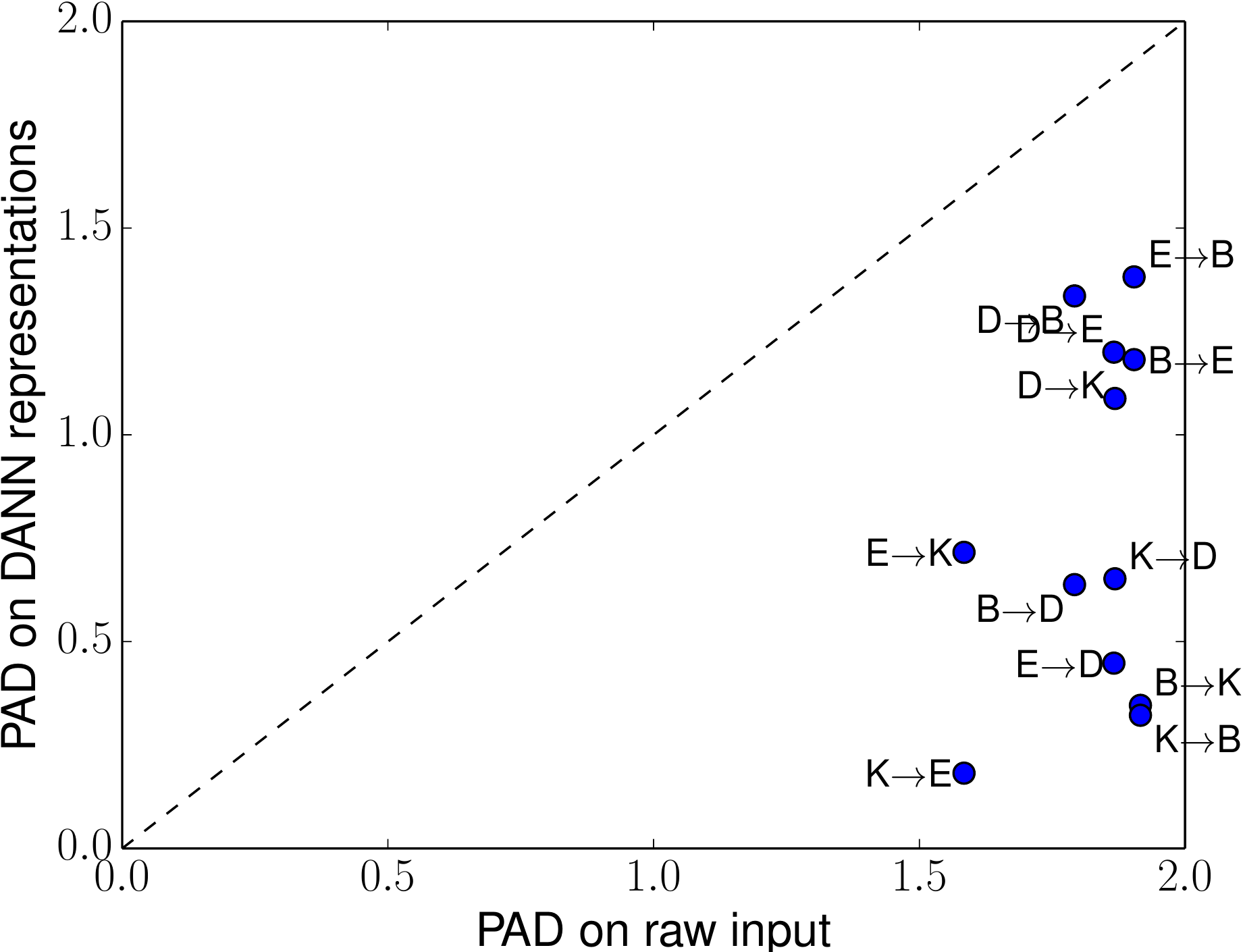}
\label{fig:PAD_a}
}\ 
\subfloat[DANN \& NN with  100 hidden neurons.]{
\includegraphics[width=.3\textwidth]{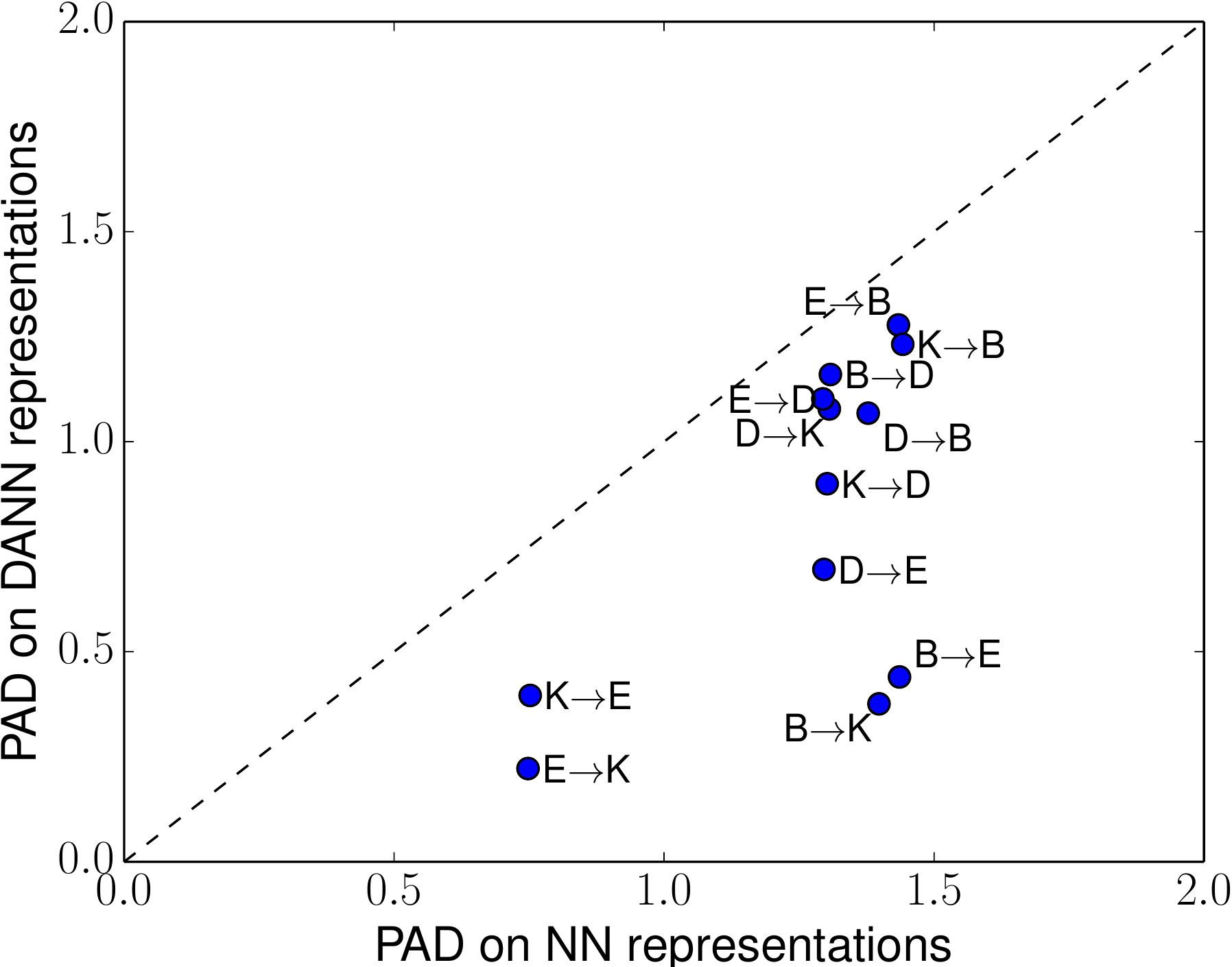}
\label{fig:PAD_b}
}\ 
\subfloat[DANN on \emph{mSDA representations.}]{
\includegraphics[width=.3\textwidth]{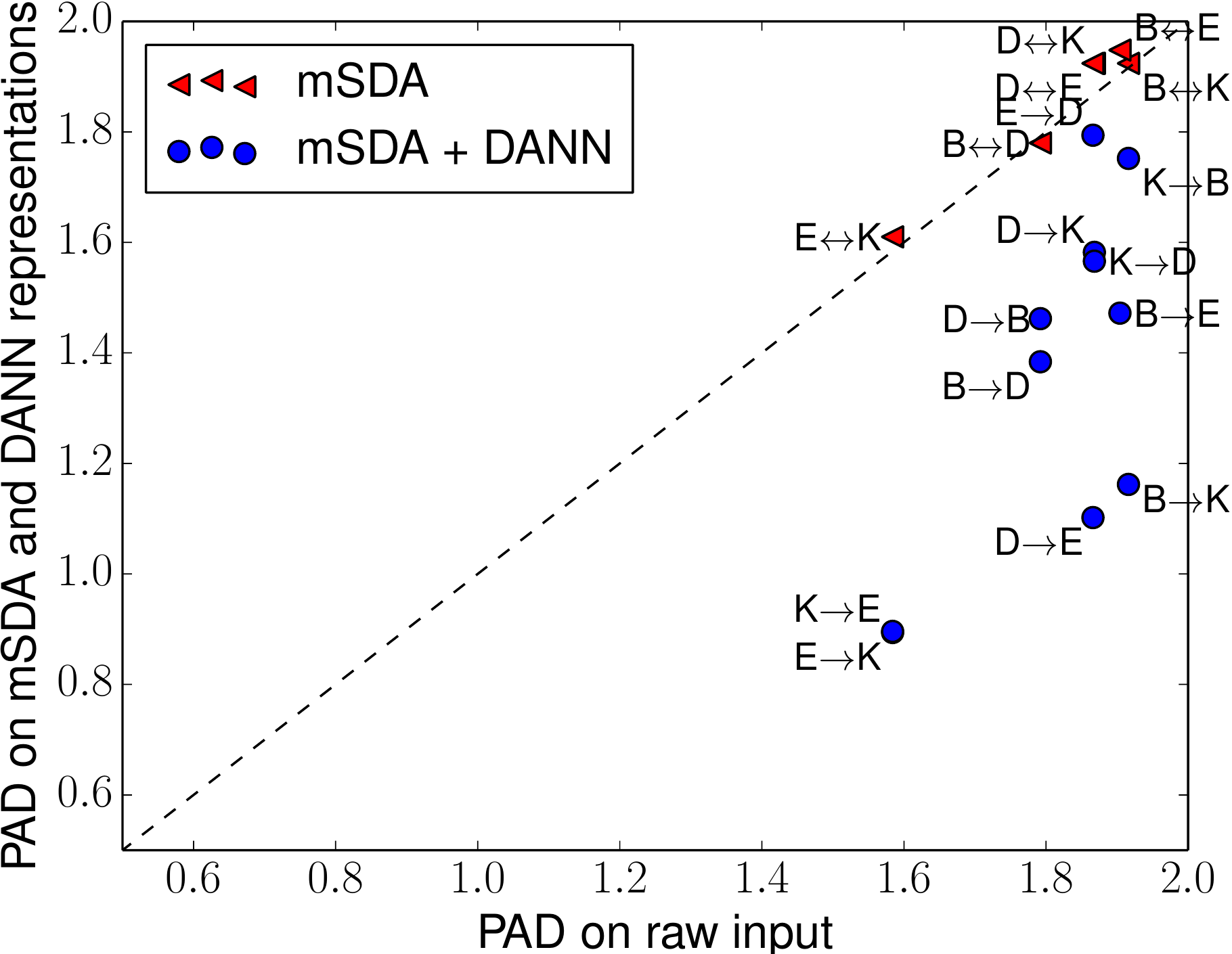}
\label{fig:PAD_c}
}
\caption{Proxy $\Acal$-distances (PAD).  Note that the PAD values of mSDA representations are symmetric when swapping source and target samples.}
\end{figure*}   

Firstly, Figure~\ref{fig:PAD_a} compares the PAD of DANN representations obtained in the experiments of Section~\ref{section:expe_original_data} (using the hyper-parameters values leading to the results of Table~\ref{table:risks}) to the PAD computed on raw data. As expected, the PAD values are driven down by the DANN representations. 

Secondly, Figure~\ref{fig:PAD_b} compares the PAD of DANN representations to the PAD of standard NN representations.
As the PAD is influenced by the hidden layer size (the discriminating power tends to increase with the representation length), we fix here the size to $100$ neurons for both algorithms. We also fix the adaptation parameter of DANN to $\lambda \simeq 0.31$; it was the value that has been selected most of the time during our preceding experiments on the \emph{Amazon Reviews} data set. Again, DANN is clearly leading to the lowest PAD values.  

Lastly, Figure~\ref{fig:PAD_c} presents two sets of results related to Section~\ref{section:autoencoders} experiments. On one hand, we reproduce the results of \citet{Chen12}, which noticed that the mSDA representations have greater PAD values than original (raw) data. Although the mSDA approach clearly helps to adapt to the target task, it seems to contradict the theory of \citeauthor{BenDavid-NIPS06}. On the other hand, we observe that, when running DANN on top of mSDA (using the hyper-parameters values leading to the results of Table~\ref{table:risks}), the obtained representations have much lower PAD values. These observations might explain the improvements provided by DANN when combined with the mSDA procedure.

\subsection{Experiments with Deep Networks on Image Classification}
\label{sect:experiments}

\def\X{{\mathbf X}}
\def\y{{\mathbf y}}

We now perform extensive evaluation of a deep version of DANN (see Subsection~\ref{section:deep_DANN}) on a number of popular image data sets and their modifications. These include large-scale data sets of small images popular with deep learning methods, and the {\sc Office} data sets \citep{Saenko10}, which are a {\em de facto} standard for domain adaptation in computer vision, but have much fewer images.

\subsubsection{Baselines} 

The following baselines are evaluated in the experiments of this subsection. The \textit{source-only} model is trained without consideration for target-domain data (no domain classifier branch included into the network). The \textit{train-on-target} model is trained on the target domain with class labels revealed. This model serves as an upper bound on DA methods, assuming that target data are abundant and the shift between the domains is considerable. 

In addition, we compare our approach against the recently proposed unsupervised DA method based on \textit{subspace alignment (SA)} \citep{Fernando13}, which is simple to setup and test on new data sets, but has also been shown to perform very well in experimental comparisons with other ``shallow'' DA methods. To boost the performance of this baseline, we pick its most important free parameter (the number of principal components) from the range $ \{ 2, \ldots, 60 \} $, so that the test performance on the target domain is maximized. To apply SA in our setting, we train a source-only model and then consider the activations of the last hidden layer in the label predictor (before the final linear classifier) as descriptors/features, and learn the mapping between the source and the target domains \citep{Fernando13}.

Since the SA baseline requires training a new classifier after adapting the features, and in order to put all the compared settings on an equal footing, we retrain the last layer of the label predictor using a standard linear SVM~\citep{liblinear} for all four considered methods (including ours; the performance on the target domain remains approximately the same after the retraining). 

For the {\sc Office} data set \citep{Saenko10}, we directly compare the performance of our full network (feature extractor and label predictor) against recent DA approaches using previously published results.

\subsubsection{CNN architectures and Training Procedure} \label{train_proc_for_classification} 

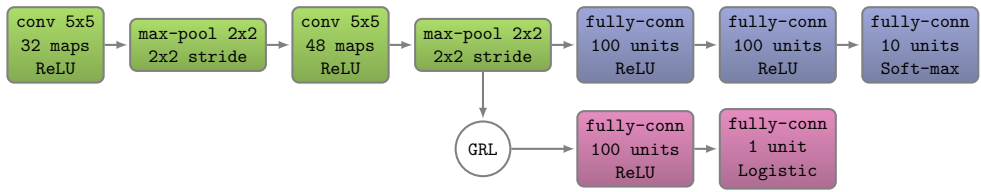
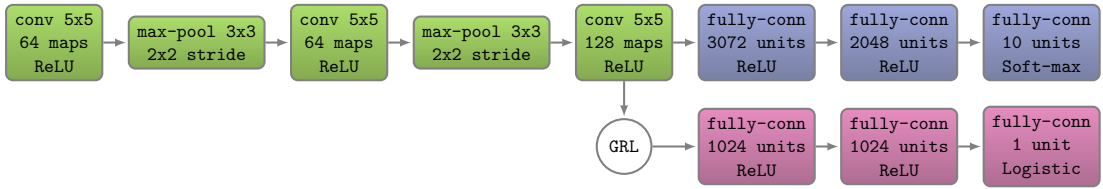
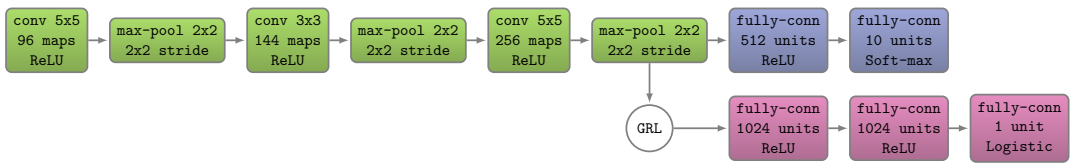
\begin{figure*}[t]
  \definecolor{fnodebottom}{RGB}{132,170,81}
  \definecolor{fnodetop}{RGB}{172,222,106}
  \definecolor{cnodebottom}{RGB}{120,128,164}
  \definecolor{cnodetop}{RGB}{158,167,218}
  \definecolor{dnodebottom}{RGB}{174,109,146}
  \definecolor{dnodetop}{RGB}{230,141,192}
  \centering
  \subfloat[MNIST architecture; inspired by the classical LeNet-5 \citep{LeCun98}.]{%
    \scalebox{0.65}{\begin{tikzpicture}[ampersand replacement=\&,
  black!50, text=black,
  node distance=4mm,
  grlnode/.style={
    align=center,
    circle,minimum size=6mm,
    inner sep=5pt,
    very thick,draw=black!50,
    font=\ttfamily
  },
  fnode/.style={
    align=center,
    rectangle,minimum size=6mm,rounded corners,
    inner sep=5pt,
    very thick,draw=black!50,
    top color=fnodetop,bottom color=fnodebottom,
    font=\ttfamily},
  cnode/.style={
    fnode,top color=cnodetop,bottom color=cnodebottom},
  dnode/.style={
    fnode,top color=dnodetop,bottom color=dnodebottom},
  vhedge/.style={
    rounded corners,to path=|- (\tikztotarget)}]
  \matrix[row sep=5mm,column sep=5mm] {
    \node (conv1) [fnode] {conv 5x5\\32 maps\\ReLU}; \&
    \node (pool1) [fnode] {max-pool 2x2\\2x2 stride}; \&
    \node (conv2) [fnode] {conv 5x5\\48 maps\\ReLU}; \&
    \node (pool2) [fnode] {max-pool 2x2\\2x2 stride}; \&
  
    \node (fc3)   [cnode] {fully-conn\\100 units\\ReLU}; \&
    \node (fc4)   [cnode] {fully-conn\\100 units\\ReLU}; \&
    \node (fc5)   [cnode] {fully-conn\\10 units\\Soft-max}; \\
  
    \& \& \&
    \node (grl) [grlnode] {GRL}; \&
  
    \node (fc1_d) [dnode] {fully-conn\\100 units\\ReLU}; \&
    \node (fc2_d) [dnode] {fully-conn\\1 unit\\Logistic}; \\
  };
  
  \path (conv1) edge[-latex,shorten >=1pt,very thick] (pool1);
  \path (pool1) edge[-latex,shorten >=1pt,very thick] (conv2);
  \path (conv2) edge[-latex,shorten >=1pt,very thick] (pool2);
  \path (pool2) edge[-latex,shorten >=1pt,very thick] (fc3);
  \path (fc3)   edge[-latex,shorten >=1pt,very thick] (fc4);
  \path (fc4)   edge[-latex,shorten >=1pt,very thick] (fc5);
  
  \path (pool2.south) edge[-latex,shorten >=1pt,very thick] (grl.north);
  \path (grl) edge[-latex,shorten >=1pt,very thick] (fc1_d);
  \path (fc1_d) edge[-latex,shorten >=1pt,very thick] (fc2_d);
\end{tikzpicture}
}}\\
  \subfloat[SVHN architecture; adopted from \citet{Srivastava14}.]{%
    \scalebox{0.65}{\begin{tikzpicture}[ampersand replacement=\&,
  black!50, text=black,
  node distance=4mm,
  grlnode/.style={
    align=center,
    circle,minimum size=6mm,
    inner sep=5pt,
    very thick,draw=black!50,
    font=\ttfamily
  },
  fnode/.style={
    align=center,
    rectangle,minimum size=6mm,rounded corners,
    inner sep=5pt,
    very thick,draw=black!50,
    top color=fnodetop,bottom color=fnodebottom,
    font=\ttfamily},
  cnode/.style={
    fnode,top color=cnodetop,bottom color=cnodebottom},
  dnode/.style={
    fnode,top color=dnodetop,bottom color=dnodebottom},
  vhedge/.style={
    rounded corners,to path=|- (\tikztotarget)}]
  \matrix[row sep=5mm,column sep=5mm] {
    \node (conv1) [fnode] {conv 5x5\\64 maps\\ReLU}; \&
    \node (pool1) [fnode] {max-pool 3x3\\2x2 stride}; \&
    \node (conv2) [fnode] {conv 5x5\\64 maps\\ReLU}; \&
    \node (pool2) [fnode] {max-pool 3x3\\2x2 stride}; \&
    \node (conv3) [fnode] {conv 5x5\\128 maps\\ReLU}; \&
  
    \node (fc4)   [cnode] {fully-conn\\3072 units\\ReLU}; \&
    \node (fc5)   [cnode] {fully-conn\\2048 units\\ReLU}; \&
    \node (fc6)   [cnode] {fully-conn\\10 units\\Soft-max}; \\
  
    \& \& \& \&
    \node (grl) [grlnode] {GRL}; \&
    \node (fc1_d) [dnode] {fully-conn\\1024 units\\ReLU}; \&
    \node (fc2_d) [dnode] {fully-conn\\1024 units\\ReLU}; \&
    \node (fc3_d) [dnode] {fully-conn\\1 unit\\Logistic}; \\
  };
  
  \path (conv1) edge[-latex,shorten >=1pt,very thick] (pool1);
  \path (pool1) edge[-latex,shorten >=1pt,very thick] (conv2);
  \path (conv2) edge[-latex,shorten >=1pt,very thick] (pool2);
  \path (pool2) edge[-latex,shorten >=1pt,very thick] (conv3);
  \path (conv3) edge[-latex,shorten >=1pt,very thick] (fc4);
  \path (fc4)   edge[-latex,shorten >=1pt,very thick] (fc5);
  \path (fc5)   edge[-latex,shorten >=1pt,very thick] (fc6);
  
  \path (conv3.south) edge[-latex,shorten >=1pt,very thick] (grl.north);
  \path (grl) edge[-latex,shorten >=1pt,very thick] (fc1_d);
  \path (fc1_d) edge[-latex,shorten >=1pt,very thick] (fc2_d);
  \path (fc2_d) edge[-latex,shorten >=1pt,very thick] (fc3_d);
\end{tikzpicture}
}}\\
  \subfloat[GTSRB architecture; we used the single-CNN baseline from \citet{Cirecsan12} as our starting point.]{%
    \scalebox{0.55}{\begin{tikzpicture}[ampersand replacement=\&,
  black!50, text=black,
  node distance=4mm,
  grlnode/.style={
    align=center,
    circle,minimum size=6mm,
    inner sep=5pt,
    very thick,draw=black!50,
    font=\ttfamily
  },
  fnode/.style={
    align=center,
    rectangle,minimum size=6mm,rounded corners,
    inner sep=5pt,
    very thick,draw=black!50,
    top color=fnodetop,bottom color=fnodebottom,
    font=\ttfamily},
  cnode/.style={
    fnode,top color=cnodetop,bottom color=cnodebottom},
  dnode/.style={
    fnode,top color=dnodetop,bottom color=dnodebottom},
  vhedge/.style={
    rounded corners,to path=|- (\tikztotarget)}]
  \matrix[row sep=5mm,column sep=5mm] {
    \node (conv1) [fnode] {conv 5x5\\96 maps\\ReLU}; \&
    \node (pool1) [fnode] {max-pool 2x2\\2x2 stride}; \&
    \node (conv2) [fnode] {conv 3x3\\144 maps\\ReLU}; \&
    \node (pool2) [fnode] {max-pool 2x2\\2x2 stride}; \&
    \node (conv3) [fnode] {conv 5x5\\256 maps\\ReLU}; \&
    \node (pool3) [fnode] {max-pool 2x2\\2x2 stride}; \&
  
    \node (fc4)   [cnode] {fully-conn\\512 units\\ReLU}; \&
    \node (fc5)   [cnode] {fully-conn\\10 units\\Soft-max}; \\
  
    \& \& \& \& \&
    \node (grl) [grlnode] {GRL}; \&
    \node (fc1_d) [dnode] {fully-conn\\1024 units\\ReLU}; \&
    \node (fc2_d) [dnode] {fully-conn\\1024 units\\ReLU}; \&
    \node (fc3_d) [dnode] {fully-conn\\1 unit\\Logistic}; \\
  };
  
  \path (conv1) edge[-latex,shorten >=1pt,very thick] (pool1);
  \path (pool1) edge[-latex,shorten >=1pt,very thick] (conv2);
  \path (conv2) edge[-latex,shorten >=1pt,very thick] (pool2);
  \path (pool2) edge[-latex,shorten >=1pt,very thick] (conv3);
  \path (conv3) edge[-latex,shorten >=1pt,very thick] (pool3);
  \path (pool3) edge[-latex,shorten >=1pt,very thick] (fc4);
  \path (fc4)   edge[-latex,shorten >=1pt,very thick] (fc5);
  
  \path (pool3.south) edge[-latex,shorten >=1pt,very thick] (grl.north);
  \path (grl) edge[-latex,shorten >=1pt,very thick] (fc1_d);
  \path (fc1_d) edge[-latex,shorten >=1pt,very thick] (fc2_d);
  \path (fc2_d) edge[-latex,shorten >=1pt,very thick] (fc3_d);
\end{tikzpicture}
}}\\
  \caption{CNN architectures used in the experiments. Boxes correspond to transformations applied to the data. Color-coding is the same as in \fig{arch}.}
  \label{fig:exper_archs}
\end{figure*}

In general, we compose feature extractor from two or three convolutional layers, picking their exact configurations from previous works. 
More precisely, four different architectures were used in our experiments. The first three are shown in \fig{exper_archs}.  For the {\sc Office} domains, we use pre-trained \texttt{AlexNet} from the \texttt{Caffe}-package \citep{Jia14}. The adaptation architecture is identical to \citet{Tzeng14}.\footnote{A 2-layer domain classifier ($x{\rightarrow}1024{\rightarrow}1024{\rightarrow}2$) is attached to the $ 256 $-dimensional bottleneck of \texttt{fc7}.}  

For the domain adaption component, we use three  ($x{\rightarrow}1024{\rightarrow}1024{\rightarrow}2$)
fully connected layers, except for {\sc MNIST} where we used a simpler ($x{\rightarrow}100{\rightarrow}2$) architecture to speed up the experiments.
Admittedly these choices for domain classifier are arbitrary, and better adaptation performance might be attained if this part of the architecture is tuned.

For the loss functions, we set $ \Lcal_y $ and $ \Lcal_d $ to be the logistic regression loss and the binomial cross-entropy respectively. Following \citet{Srivastava14} we also use dropout and $ \ell_2 $-norm restriction when we train the SVHN architecture.

The other hyper-parameters are not selected through a grid search as in the small scale experiments of Section~\ref{section:experiments_shallow}, which would be computationally costly. 
Instead, the learning rate is adjusted during the stochastic gradient descent  using the following formula:
\begin{equation*}
  \mu_p = \frac{\mu_0}{(1 + \alpha \cdot p)^\beta} \, , 
\end{equation*}
where $ p $ is the training progress linearly changing from 0 to 1, $ \mu_0 = 0.01 $, $ \alpha = 10 $ and $ \beta = 0.75 $ (the schedule was optimized to promote convergence and low error on the \emph{source} domain). A momentum term of $0.9$ is also used.

The domain adaptation parameter $\lambda$ is initiated at $0$ and is gradually changed  to $1$ using the following schedule:
\begin{equation*}
  \lambda_p \ =\ \frac{2}{1 + \exp(-\gamma \cdot p)} - 1\,,
\end{equation*}
where $\gamma$ was set to $10$ in all experiments (the schedule was not optimized/tweaked). 
This strategy allows the domain classifier to be less sensitive to noisy signal at the early stages of the training procedure.
Note however that these $\lambda_p$ were used only for updating the \emph{feature extractor} component $G_f$. For updating the \emph{domain classification} component, we used a fixed $\lambda=1$, to ensure that the latter trains as fast as the \emph{label predictor} $G_y$.\footnote{Equivalently, one can use the same $\lambda_p$ for both feature extractor and domain classification components, but use a learning rate of $\mu/\lambda_p$ for the latter.}

Finally, note that the model is trained on $128$-sized batches (images are preprocessed by the mean subtraction). A half of each batch is populated by the samples from the source domain (with known labels), the rest constitutes the target domain (with labels not revealed to the algorithms except for the train-on-target baseline).

\subsubsection{Visualizations}

\begin{figure*}[t]
  \centering
  \small{{\sc MNIST $ \rightarrow $ MNIST-M}: top feature extractor layer}\\
  \setcounter{subfigure}{0}
  \subfloat[Non-adapted]{%
    \includegraphics[width=0.4\textwidth]{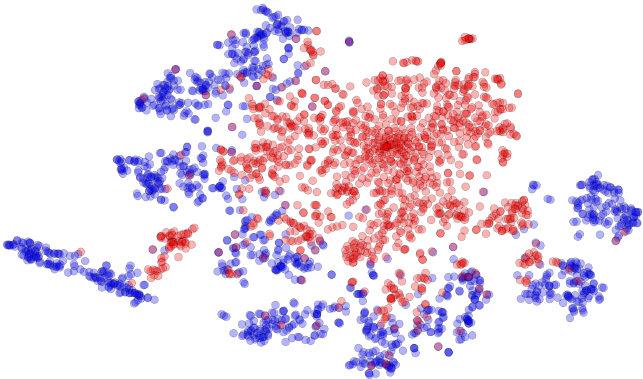}}\hfill%
  \subfloat[Adapted]{%
    \includegraphics[width=0.4\textwidth]{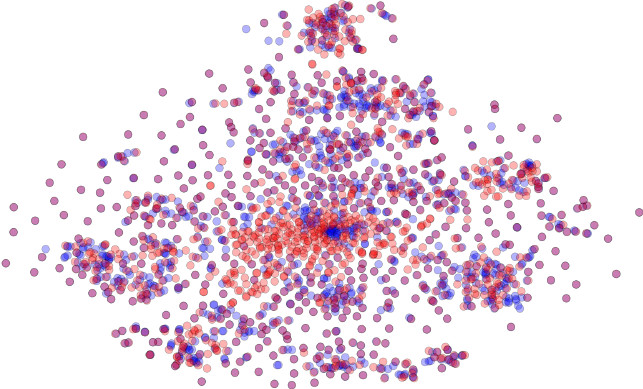}}%

\medskip
  \small{{\sc Syn Numbers $ \rightarrow $ SVHN}: last hidden layer of the label predictor}\\
  \setcounter{subfigure}{0}
  \subfloat[Non-adapted]{%
    \includegraphics[width=0.4\textwidth]{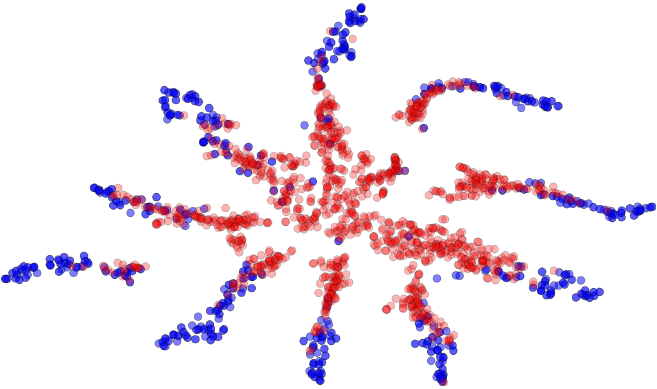}}\hfill%
  \subfloat[Adapted]{%
    \includegraphics[width=0.4\textwidth]{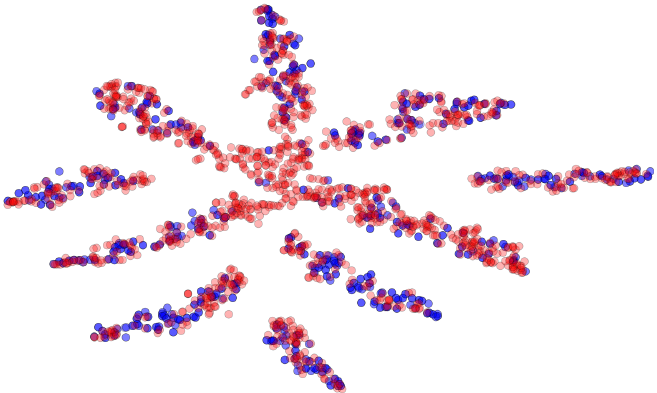}}%
  \caption{The effect of adaptation on the distribution of the extracted features (best viewed in color). The figure shows t-SNE \citep{Maaten13} visualizations of the CNN's activations {\bf (a)} in case when no adaptation was performed and {\bf (b)} in case when our adaptation procedure was incorporated into training. {\it Blue} points correspond to the source domain examples, while {\it red} ones correspond to the target domain. In all cases, the adaptation in our method makes the two distributions of features much closer.}
  \label{fig:exper_adapt_vis}
\end{figure*}

We use t-SNE \citep{Maaten13} projection to visualize feature distributions at different points of the network, while color-coding the domains (\fig{exper_adapt_vis}). 
As we already observed with the shallow version of DANN (see Figure~\ref{fig:2moons}),
there is a strong correspondence between the success of the adaptation in terms of the classification accuracy for the target domain, and the overlap between the domain distributions in such visualizations.

\subsubsection{Results On Image Data Sets}
\label{sect:exper_quant}
\renewcommand{\example}[1]{\raisebox{-.4\height}{\includegraphics[width=\figurewidth]{figures/domains_examples/#1}}}

\afterpage{
\begin{figure*}
  \centering
  \setlength{\tabcolsep}{0pt}
  \setlength\figurewidth{0.05\textwidth}
\begin{sc}
  \begin{small}
  \begin{tabular}{r@{\hskip 0.5cm} ccc c@{\hskip 0.4cm} ccc c@{\hskip 0.4cm} ccc c@{\hskip 0.4cm} ccc}
    &
    \multicolumn{3}{c}{MNIST} & &
    \multicolumn{3}{c}{Syn Numbers} & &
    \multicolumn{3}{c}{SVHN} & &
    \multicolumn{3}{c}{Syn Signs}\\
    
    Source &
    \example{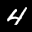} &
    \example{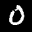} &
    \example{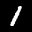} & &
    
    \example{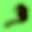} &
    \example{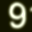} &
    \example{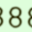} & &
    
    \example{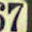} &
    \example{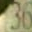} &
    \example{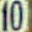} & &
    
    \example{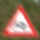} &
    \example{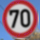} &
    \example{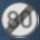}\\
    
    Target &
    \example{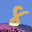} &
    \example{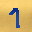} &
    \example{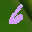} & &
    
    \example{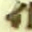} &
    \example{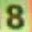} &
    \example{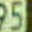} & &
    
    \example{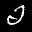} &
    \example{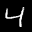} &
    \example{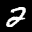} & &
    
    \example{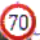} &
    \example{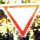} &
    \example{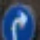}\\
    
    &
    \multicolumn{3}{c}{\rule{0pt}{0.35cm} MNIST-M} & &
    \multicolumn{3}{c}{SVHN} & &
    \multicolumn{3}{c}{MNIST} & &
    \multicolumn{3}{c}{GTSRB}\\
  \end{tabular}
  \end{small}
  \end{sc}
  \caption{Examples of domain pairs used in the experiments. See \sect{exper_quant} for details.}
  \label{fig:exper_domains_examples}
\end{figure*}
\begin{table*}[t]
\centering
    \begin{small}
      \begin{sc}
        \renewcommand{\arraystretch}{1.3}
        \rowcolors{2}{black!10}{}
        \begin{tabular}{l r | c c c c}
          \hline
          \multirow{2}{*}{Method} & {\scriptsize Source} & MNIST & Syn Numbers & SVHN & Syn Signs \\
          & {\scriptsize Target} & MNIST-M & SVHN & MNIST & GTSRB \\
          \hline
          \multicolumn{2}{l |}{Source only} & 
          $ .5225 $                      & $ .8674 $                      & $ .5490 $                      & $ .7900 $                      \\
          \multicolumn{2}{l |}{SA {\rm \citep{Fernando13}}} & 
          $ .5690 \; (4.1\%) $           & $ .8644 \; (-5.5\%) $          & $ .5932 \; (9.9\%) $           & $ .8165 \; (12.7\%) $          \\
          \multicolumn{2}{l |}{DANN} & 
          $ \mathbf{.7666} \; (52.9\%) $ & $ \mathbf{.9109} \; (79.7\%) $ & $ \mathbf{.7385} \; (42.6\%) $ & $ \mathbf{.8865} \; (46.4\%) $ \\
          \multicolumn{2}{l |}{Train on target} & 
          $ .9596 $                      & $ .9220 $                      & $ .9942 $                      & $ .9980 $                      \\
          \hline
        \end{tabular}
      \end{sc}
    \end{small}
    \caption{Classification accuracies for digit image classifications for different source and target domains. {\sc MNIST-M} corresponds to difference-blended digits over non-uniform background. The first row corresponds to the lower performance bound (\ie, if no adaptation is performed). The last row corresponds to training on the target domain data with known class labels (upper bound on the DA performance). For each of the two DA methods \citep[ours and][]{Fernando13} we show how much of the gap between the lower and the upper bounds was covered (in brackets). For all five cases, our approach outperforms \citet{Fernando13} considerably, and covers a big portion of the gap.}
  \label{tab:results}
\end{table*}
\begin{table*} %
\centering
    \begin{small}
      \begin{sc}
        \renewcommand{\arraystretch}{1.3}
        \rowcolors{2}{black!10}{}
        \begin{tabular}{l r | c c c}
          \hline
          \multirow{2}{*}{Method} & {\scriptsize Source} & Amazon & DSLR & Webcam \\
          & {\scriptsize Target} & Webcam & Webcam & DSLR \\
          \hline
          \multicolumn{2}{l |}{GFK(PLS, PCA) {\rm \citep{Gong12}}} & 
          $ .197  $ & $ .497 $ & $ .6631 $\\ 
          \multicolumn{2}{l |}{SA* {\rm \citep{Fernando13}}} & 
          $ .450 $ & $ .648 $ & $ .699 $\\ 
          \multicolumn{2}{l |}{DLID {\rm \citep{Chopra13}}} & 
          $ .519 $ & $ .782 $ & $ .899 $\\
          \multicolumn{2}{l |}{DDC {\rm \citep{Tzeng14}}} & 
          $ .618 $ & $ .950 $ & $ .985 $\\
          \multicolumn{2}{l |}{DAN {\rm \citep{Long15}}} & 
          $ .685 $ & $ .960 $ & $ .990  $\\ 
          \hline
          \multicolumn{2}{l |}{Source only} & 
          $ .642 $ & $ .961 $ & $ .978 $\\ 
          \multicolumn{2}{l |}{DANN} & 
          $ \mathbf{ .730 } $ & $ \mathbf{ .964 } $ & $ \mathbf{ .992 } $\\
          \hline
        \end{tabular}
      \end{sc}
    \end{small}
    \caption{Accuracy evaluation of different DA approaches on the standard {\sc Office} \citep{Saenko10} data set. All methods (except SA) are evaluated in the ``fully-transductive'' protocol (some results are reproduced from \citealp{Long15}). Our method (last row) outperforms competitors setting the new state-of-the-art.}
  \label{tab:results_office}
\end{table*}
\clearpage
}

We now discuss the experimental settings and the results. In each case, we train on the source data set and test on a different target domain data set, with considerable shifts between domains (see \fig{exper_domains_examples}). The results are summarized in \tab{results} and \tab{results_office}. 

\vspace{2mm}\noindent {\it MNIST $ \rightarrow $ MNIST-M.}
Our first experiment deals with the MNIST data set~\citep{LeCun98} (source). In order to obtain the target domain ({\sc MNIST-M}) we blend digits from the original set over patches randomly extracted from color photos from BSDS500 \citep{Arbelaez11}. This operation is formally defined for two images $ I^{1}, I^{2} $ as $ I_{ijk}^{out} = | I_{ijk}^{1} - I_{ijk}^{2} | $, where $ i, j $ are the coordinates of a pixel and $ k $ is a channel index. In other words, an output sample is produced by taking a patch from a photo and inverting its pixels at positions corresponding to the pixels of a digit. For a human the classification task becomes only slightly harder compared to the original data set (the digits are still clearly distinguishable) whereas for a CNN trained on MNIST this domain is quite distinct, as the background and the strokes are no longer constant. Consequently, the source-only model performs poorly. Our approach succeeded at aligning feature distributions (\fig{exper_adapt_vis}), which led to successful adaptation results (considering that the adaptation is unsupervised). At the same time, the improvement over source-only model achieved by subspace alignment (SA) \citep{Fernando13} is quite modest, thus highlighting the difficulty of the adaptation task. 

\vspace{2mm}\noindent {\it Synthetic numbers $ \rightarrow $ SVHN.}
To address a common scenario of training on synthetic data and testing on  real data, we use Street-View House Number data set {\sc SVHN} \citep{Netzer11} as the target domain and synthetic digits as the source. The latter ({\sc Syn ~Numbers}) consists of $\approx \num[group-separator={,}]{500000}$ images generated by ourselves from Windows\textsuperscript{\tiny TM} fonts by varying the text (that includes different one-, two-, and three-digit numbers), positioning, orientation, background and stroke colors, and the amount of blur. The degrees of variation were chosen manually to simulate {\sc SVHN}, however the two data sets are still rather distinct, the biggest difference being the structured clutter in the background of {\sc SVHN} images. 

The proposed backpropagation-based technique works well covering almost 80\% of the gap between training with source data only and training on target domain data with known target labels. In contrast, SA~\citep{Fernando13} results in a slight classification accuracy drop (probably due to the information loss during the dimensionality reduction), indicating that the adaptation task is even more challenging than in the case of the {\sc MNIST} experiment.

\vspace{2mm}\noindent {\it MNIST $ \leftrightarrow $ SVHN.}
In this experiment, we further increase the gap between distributions, and test on {\sc MNIST} and {\sc SVHN}, which are significantly different in appearance. Training on SVHN even without adaptation is challenging --- classification error stays high during the first 150 epochs. In order to avoid ending up in a poor local minimum we, therefore, do not use learning rate annealing here. Obviously, the two directions ({\sc MNIST} $ \rightarrow $ {\sc SVHN} and {\sc SVHN} $ \rightarrow $ {\sc MNIST}) are not equally difficult. As {\sc SVHN} is more diverse, a model trained on SVHN is expected to be more generic and to perform reasonably on the MNIST data set. This, indeed, turns out to be the case and is supported by the appearance of the feature distributions. We observe a quite strong separation between the domains when we feed them into the CNN trained solely on {\sc MNIST}, whereas for the {\sc SVHN}-trained network the features are much more intermixed. This difference probably explains why our method succeeded in improving the performance by adaptation in the {\sc SVHN} $ \rightarrow $ {\sc MNIST} scenario (see \tab{results}) but not in the opposite direction (SA is not able to perform adaptation in this case either). Unsupervised adaptation from {\sc MNIST} to {\sc SVHN} gives a failure example for our approach: it doesn't manage to improve upon the performance of the non-adapted model which achieves $\approx 0.25$ accuracy (we are unaware of any unsupervised DA methods capable of performing such adaptation).

\vspace{2mm}\noindent {\it Synthetic Signs $ \rightarrow $ GTSRB.}
Overall, this setting is similar to the {\sc Syn Numbers} $ \rightarrow $ {\sc SVHN} experiment, except the distribution of the features is more complex due to the significantly larger number of classes (43 instead of 10). For the source domain we obtained $ \num[group-separator={,}]{100000} $ synthetic images (which we call {\sc Syn~Signs}) simulating various imaging conditions. In the target domain, we use $ \num[group-separator={,}]{31367} $ random training samples for unsupervised adaptation and the rest for evaluation. Once again, our method achieves a sensible increase in performance proving its suitability for the synthetic-to-real data adaptation.

\begin{figure}
  \centering
  \setlength\figureheight{4cm}
  \setlength\figurewidth{6.8cm}
\definecolor{mycolor1}{rgb}{0.00000,0.00000,0.00000}%
\definecolor{mycolor2}{rgb}{0.00000,1.00000,0.40000}%
\definecolor{mycolor3}{rgb}{0.00000,0.40000,1.00000}%
\definecolor{mycolor4}{rgb}{0.80000,0.00000,1.00000}%
\begin{tikzpicture}[ampersand replacement=\&,font=\scriptsize]

\begin{axis}[%
width=0.984128\figurewidth,
height=\figureheight,
at={(0\figurewidth,0\figureheight)},
scale only axis,
xmin=0,
xmax=500000,
xlabel={Batches seen},
ymin=0.06,
ymax=0.22,
ylabel={Validation error},
ytick={0.05,0.1,0.15,0.2},
axis x line*=bottom,
axis y line*=left,
reverse legend,
legend style={at={($ (1,1) + (-0.1cm,-0.1cm) $)},anchor=north east,legend cell align=left,align=left,draw=black,fill=white,fill opacity=0.8,draw opacity=1.0,text opacity=1.0}
]
\addplot [color=red,solid,line width=1.0pt]
  table[row sep=crcr]{%
10000	0.08684\\
20000	0.076639\\
30000	0.079061\\
40000	0.077404\\
50000	0.075491\\
60000	0.075363\\
70000	0.073578\\
80000	0.07243\\
90000	0.071538\\
100000	0.071283\\
110000	0.068095\\
120000	0.067712\\
130000	0.06886\\
140000	0.068222\\
150000	0.065672\\
160000	0.066055\\
170000	0.066182\\
180000	0.064652\\
190000	0.065544\\
200000	0.064779\\
210000	0.065034\\
220000	0.065034\\
230000	0.065034\\
240000	0.066182\\
250000	0.0651620000000001\\
260000	0.064397\\
270000	0.065034\\
280000	0.065289\\
290000	0.065289\\
300000	0.06631\\
310000	0.064907\\
320000	0.064397\\
330000	0.066055\\
340000	0.064779\\
350000	0.0658\\
360000	0.0651620000000001\\
370000	0.064779\\
380000	0.064907\\
390000	0.0654169999999999\\
400000	0.064524\\
410000	0.064397\\
420000	0.064779\\
430000	0.064524\\
440000	0.064524\\
450000	0.063504\\
460000	0.065544\\
470000	0.064779\\
480000	0.064014\\
490000	0.064907\\
500000	0.0654169999999999\\
};
\addlegendentry{Syn + Real Adapted};

\addplot [color=mycolor1,solid,line width=1.0pt]
  table[row sep=crcr]{%
10000	0.076511\\
20000	0.075873\\
30000	0.0748529999999999\\
40000	0.075491\\
50000	0.0742159999999999\\
60000	0.0739610000000001\\
70000	0.074726\\
80000	0.073451\\
90000	0.073578\\
100000	0.073578\\
110000	0.073068\\
120000	0.074343\\
130000	0.073578\\
140000	0.071793\\
150000	0.0745980000000001\\
160000	0.073196\\
170000	0.0745980000000001\\
180000	0.074471\\
190000	0.0742159999999999\\
200000	0.073706\\
210000	0.074343\\
220000	0.0739610000000001\\
230000	0.073196\\
240000	0.073706\\
250000	0.073196\\
260000	0.072813\\
270000	0.072558\\
280000	0.073833\\
290000	0.07243\\
300000	0.074088\\
310000	0.0739610000000001\\
320000	0.072813\\
330000	0.072303\\
340000	0.073451\\
350000	0.073833\\
360000	0.072558\\
370000	0.073323\\
380000	0.073196\\
390000	0.073068\\
400000	0.072686\\
410000	0.073068\\
420000	0.073068\\
430000	0.073323\\
440000	0.072686\\
450000	0.073578\\
460000	0.073833\\
470000	0.072686\\
480000	0.072941\\
490000	0.072686\\
500000	0.073451\\
};
\addlegendentry{Syn + Real};

\addplot [color=mycolor2,solid,line width=1.0pt]
  table[row sep=crcr]{%
10000	0.214869\\
20000	0.189748\\
30000	0.167559\\
40000	0.157613\\
50000	0.152767\\
60000	0.150854\\
70000	0.143841\\
80000	0.143586\\
90000	0.13976\\
100000	0.137847\\
110000	0.135425\\
120000	0.13721\\
130000	0.135042\\
140000	0.131854\\
150000	0.130069\\
160000	0.130706\\
170000	0.126626\\
180000	0.125988\\
190000	0.125733\\
200000	0.127008\\
210000	0.125861\\
220000	0.125988\\
230000	0.125733\\
240000	0.125606\\
250000	0.125096\\
260000	0.124586\\
270000	0.122673\\
280000	0.122928\\
290000	0.122035\\
300000	0.12229\\
310000	0.121398\\
320000	0.12025\\
330000	0.121653\\
340000	0.12076\\
350000	0.119867\\
360000	0.120632\\
370000	0.118975\\
380000	0.11872\\
390000	0.117955\\
400000	0.119357\\
410000	0.11872\\
420000	0.118975\\
430000	0.11923\\
440000	0.118592\\
450000	0.12025\\
460000	0.119612\\
470000	0.119357\\
480000	0.119102\\
490000	0.11821\\
500000	0.116934\\
};
\addlegendentry{Syn Adapted};

\addplot [color=mycolor3,solid,line width=1.0pt]
  table[row sep=crcr]{%
10000	0.215889\\
20000	0.20352\\
30000	0.194338\\
40000	0.196251\\
50000	0.198801\\
60000	0.193318\\
70000	0.193318\\
80000	0.198419\\
90000	0.193191\\
100000	0.191915\\
110000	0.191405\\
120000	0.189492\\
130000	0.188855\\
140000	0.18656\\
150000	0.191915\\
160000	0.19268\\
170000	0.18911\\
180000	0.185922\\
190000	0.184392\\
200000	0.189492\\
210000	0.185539\\
220000	0.188855\\
230000	0.186049\\
240000	0.184137\\
250000	0.18656\\
260000	0.187835\\
270000	0.186942\\
280000	0.184264\\
290000	0.188472\\
300000	0.185539\\
310000	0.184519\\
320000	0.186815\\
330000	0.186177\\
340000	0.183754\\
350000	0.183627\\
360000	0.183882\\
370000	0.185284\\
380000	0.183244\\
390000	0.186177\\
400000	0.185539\\
410000	0.185029\\
420000	0.184519\\
430000	0.184137\\
440000	0.185157\\
450000	0.186177\\
460000	0.183372\\
470000	0.184647\\
480000	0.185922\\
490000	0.185157\\
500000	0.188217\\
};
\addlegendentry{Syn};

\addplot [color=mycolor4,solid,line width=1.0pt]
  table[row sep=crcr]{%
10000	0.209768\\
20000	0.207473\\
30000	0.206197\\
40000	0.204157\\
50000	0.204157\\
60000	0.202117\\
70000	0.203902\\
80000	0.202499\\
90000	0.202627\\
100000	0.202627\\
110000	0.202117\\
120000	0.203009\\
130000	0.202754\\
140000	0.202244\\
150000	0.201734\\
160000	0.202499\\
170000	0.200842\\
180000	0.201479\\
190000	0.202117\\
200000	0.202499\\
210000	0.201607\\
220000	0.201862\\
230000	0.202244\\
240000	0.202372\\
250000	0.202372\\
260000	0.201989\\
270000	0.20352\\
280000	0.203009\\
290000	0.202372\\
300000	0.204285\\
310000	0.204795\\
320000	0.204157\\
330000	0.203137\\
340000	0.202882\\
350000	0.202882\\
360000	0.203647\\
370000	0.203009\\
380000	0.20352\\
390000	0.203264\\
400000	0.203902\\
410000	0.202627\\
420000	0.202372\\
430000	0.202627\\
440000	0.203775\\
450000	0.203392\\
460000	0.203264\\
470000	0.203392\\
480000	0.203009\\
490000	0.203392\\
500000	0.20403\\
};
\addlegendentry{Real};

\end{axis}
\end{tikzpicture}%
  \caption{Results for the traffic signs classification in the semi-supervised setting. {\it Syn} and {\it Real} denote available labeled data ($ \num[group-separator={,}]{100000} $ synthetic and $430$ real images respectively); {\it Adapted} means that $\approx \num[group-separator={,}]{31000}$ unlabeled target domain images were used for adaptation. The best performance is achieved by employing both the labeled samples and the large unlabeled corpus in the target domain.}
  \label{fig:exper_semi_test}
\end{figure}
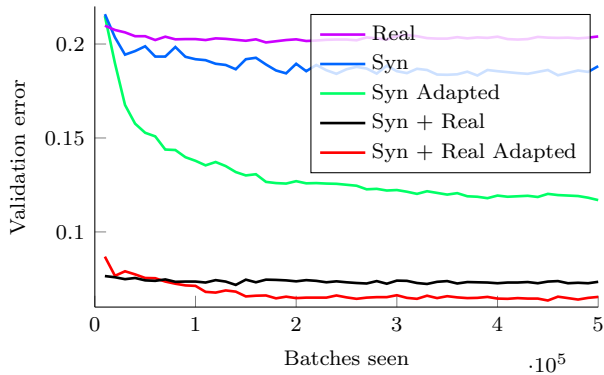

As an additional experiment, we also evaluate the proposed algorithm for semi-supervised domain adaptation, \ie, when one is additionally provided with a small amount of labeled target data. Here, we reveal $430$ labeled examples (10 samples per class) and add them to the training set for the label predictor. \fig{exper_semi_test} shows the change of the validation error throughout the training. While the graph clearly suggests that our method can be beneficial in the semi-supervised setting, thorough verification of semi-supervised setting is left for future work.

\vspace{2mm}\noindent {\it Office data set.} 
We finally evaluate our method on {\sc Office} data set, which is a collection of three distinct domains: {\sc Amazon}, {\sc DSLR}, and {\sc Webcam}. Unlike previously discussed data sets, {\sc Office} is rather small-scale with only 2817 labeled images spread across 31 different categories in the largest domain. The amount of available data is crucial for a successful training of a deep model, hence we opted for the fine-tuning of the CNN pre-trained on the ImageNet (\texttt{AlexNet} from the \texttt{Caffe} package, see \citealp{Jia14}) as it is done in some recent DA works \citep{Donahue14,Tzeng14,Hoffman14,Long15}. We make our approach more comparable with \citet{Tzeng14} by using exactly the same network architecture replacing domain mean-based regularization with the domain classifier.

Following previous works, we assess the performance of our method across three transfer tasks most commonly used for evaluation. Our training protocol is adopted from \citet{Gong13,Chopra13,Long15} as during adaptation we use all available labeled source examples and unlabeled target examples (the premise of our method is the abundance of unlabeled data in the target domain). Also, all source domain data are used for training. Under this ``fully-transductive'' setting, our method is able to improve previously-reported state-of-the-art accuracy for unsupervised adaptation very considerably (\tab{results_office}), especially in the most challenging {\sc Amazon} $ \rightarrow $ {\sc Webcam} scenario (the two domains with the largest domain shift). 

Interestingly, in all three experiments we observe a slight over-fitting (performance on the target domain degrades while accuracy on the source continues to improve) as training progresses, however, it doesn't ruin the validation accuracy. Moreover, switching off the domain classifier branch makes this effect far more apparent, from which we conclude that our technique serves as a regularizer.

\subsection{Experiments with Deep Image Descriptors for Re-Identification}
\label{section:deep_image}

In this section we discuss the application of the described adaptation method to person re-identification \textit{(re-id}) problem.  The task of person re-identification is to associate people seen from different camera views. More formally, it can be defined as follows: given two sets of images from different cameras (\textit{probe} and \textit{gallery}) such that each person depicted in the probe set has an image in the gallery set,  for each image of a person from the probe set find an image of the same person in the gallery set.  Disjoint camera views, different illumination conditions, various poses and low quality of data make this problem difficult  even for humans (\eg, \citealp{LiuLGW13}, reports human performance at Rank1=$71.08\%$).  

Unlike classification problems that are discussed above, re-identification problem implies that each image is mapped to a vector descriptor. The distance between descriptors is then used to match images from the probe set and the gallery set.
To evaluate results of re-id methods the \textit{Cumulative Match Characteristic} (CMC) curve is commonly used. It is a plot of the identification rate (recall) at rank-$k$, that is the probability of the matching gallery image to be within the closest $k$ images (in terms of descriptor distance) to the probe image.

Most existing works train descriptor mappings and evaluate them within the same data set containing images from a certain camera network with similar imaging conditions. Several papers, however, observed that the performance of the resulting re-identification systems drops very considerably when descriptors trained on one data set and tested on another. It is therefore natural to handle such cross-domain evaluation as a domain-adaptation problem, where each camera network (data set) constitutes a domain.

Recently, several papers  with significantly improved re-identification performance \citep{ZhangS14a,ZhaoOW14,Paisitkriangkrai15} have been presented, with \citet{MaLYL15} reporting good results in cross-data-set evaluation scenario. At the moment, deep learning methods \citep{YiLL14} do not achieve state-of-the-art results probably because of the limited size of the training sets. Domain adaptation thus represents a viable direction for improving deep re-identification descriptors.

\subsubsection{Data Sets and Protocols} 

Following \citet{MaLYL15}, we use PRID \citep{Hirzer_h.:person}, VIPeR \citep{Gray07evaluatingappearance}, CUHK \citep{LiW13} as target data sets for our experiments.  The \textit{PRID} data set exists in two versions, and as in \citet{MaLYL15} we use a single-shot variant. It contains images of $385$ persons viewed from camera A and images of $749$ persons viewed from camera B,  $200$ persons appear in both cameras. The \textit{VIPeR} data set also contains images taken with two cameras, and in total $632$ persons are captured, for every person there is one image for each of the two camera views. The \textit{CUHK} data set consists of images from five pairs of cameras, two images for each person from each of the two cameras. We refer to the subset of this data set that includes the first pair of cameras only as \textit{CUHK/p1} (as most papers use this subset).
See \fig{reidsamples} for samples of these data sets.

\newlength\reidheight
\setlength{\reidheight}{2.5cm}

\addtolength{\tabcolsep}{-3pt}

\begin{figure*}
\centering
\begin{tabular}{cccc|cccc|cccc}
\includegraphics[height=\reidheight]{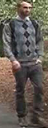}&
\includegraphics[height=\reidheight]{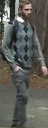}&
\includegraphics[height=\reidheight]{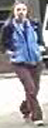}&
\includegraphics[height=\reidheight]{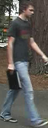}&
\includegraphics[height=\reidheight]{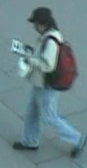}&
\includegraphics[height=\reidheight]{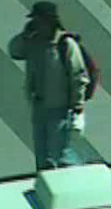}&
\includegraphics[height=\reidheight]{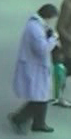}&
\includegraphics[height=\reidheight]{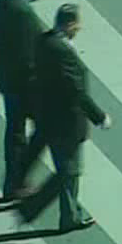}&
\includegraphics[height=\reidheight]{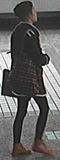}&
\includegraphics[height=\reidheight]{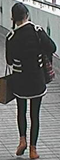}&
\includegraphics[height=\reidheight]{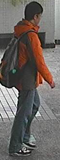}&
\includegraphics[height=\reidheight]{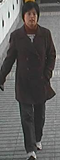}\\
\multicolumn{4}{c}{VIPER}&
\multicolumn{4}{c}{PRID}&
\multicolumn{4}{c}{CUHK}
\end{tabular}
\caption{Matching and non-matching pairs of probe-gallery images from different person re-identification data sets. The three data sets are treated as different domains in our experiments.}
\label{fig:reidsamples}
\end{figure*}

\addtolength{\tabcolsep}{3pt}

We perform extensive experiments for various pairs of data sets, where one data set serves as a source domain, \ie, it is used to train a descriptor mapping in a supervised way with known correspondences between probe and gallery images. The second data set is used as a target domain, so that images from that data set are used without probe-gallery correspondence.

In more detail, CUHK/p1 is used for experiments when CUHK serves as a target domain and two settings (``whole CUHK'' and CUHK/p1) are used for experiments when CUHK serves as a source domain. Given PRID as a target data set, we randomly choose 100 persons appearing in both camera views as training set. The images of the other 100 persons from camera A are used as probe, all images from camera B excluding those used in training (649 in total) are used as gallery at test time. For VIPeR, we use random 316 persons for training and all others for testing. For CUHK, 971 persons are split into 485 for training and 486 for testing.
Unlike \citet{MaLYL15}, we use all images in the first  pair of cameras of CUHK instead of choosing one image of a person from each camera view. We also performed two experiments with all images of the whole CUHK data set as source domain and VIPeR and PRID data sets as target domains as in the  original paper \citep{YiLL14}.

Following \citet{YiLL14}, we augmented our data with mirror images, and during test time we calculate similarity score between two images as the mean of the four scores corresponding to different flips of the two compared images. In case of CUHK, where there are 4 images (including mirror images) for each of the two camera views for each person, all 16 combinations' scores are averaged. 

\subsubsection{CNN architectures and Training Procedure} 

In our experiments, we use siamese architecture described in \citet{YiLL14} (\textit{Deep Metric Learning} or \textit{DML}) for learning deep image descriptors on the source data set.
This architecture incorporates two convolution layers (with $7\times7$ and $5\times5$ filter banks), followed by ReLU and max pooling, and one fully-connected layer, which gives $500$-dimensional descriptors as an output. There are three parallel flows within the CNN for processing three part of an image: the upper, the middle, and the lower one. The first convolution layer shares parameters between three parts, and the outputs of the second convolution layers are concatenated.
During training, we follow \citet{YiLL14} and calculate pairwise cosine similarities between $500$-dimensional features within each batch and backpropagate the loss for all pairs within batch. 

To perform domain-adversarial training, we construct a DANN architecture.  The feature extractor includes the two convolutional layers (followed by max-pooling and ReLU) discussed above. The label predictor in this case is replaced with \textit{descriptor predictor} that includes one fully-connected layer. The domain classifier includes two fully-connected layers with $500$ units in the intermediate representation ($x{\rightarrow}500{\rightarrow}1$). 

For the verification loss function in the descriptor predictor we used Binomial Deviance loss, defined in \citet{YiLL14} with similar parameters: $\alpha = 2$, $\beta = 0.5$, $c = 2$ (the asymmetric cost parameter for negative pairs). The domain classifier is trained with logistic loss as in subsection  \ref{train_proc_for_classification}.

We used learning rate fixed to $0.001$ and momentum of $0.9$. The schedule of adaptation similar to the one described in subsection \ref{train_proc_for_classification} was used. We also inserted dropout layer with rate $0.5$ after the concatenation of outputs of the second max-pooling layer. $128$-sized batches were used for source data and $128$-sized batches for target data. 

\subsubsection{Results on Re-identification data sets} 

Figure \ref{fig:adaptresults} shows results in the form of CMC-curves for eight pairs of data sets. Depending on the hardness of the annotation problem we trained either for 50,000 iterations (CUHK/p1 $\rightarrow$ VIPeR, VIPeR $\rightarrow$ CUHK/p1, PRID $\rightarrow$ VIPeR) or for 20,000 iterations (the other five pairs). 

\afterpage{
\begin{figure}[t]
\centering
  \begin{tabular}{p{5cm}  p{5cm}  p{5cm} }
      
        \setlength\figureheight{3.5cm}
        \setlength\figurewidth{3.5cm}
        \begin{tikzpicture}[ampersand replacement=\&,font=\scriptsize]

\begin{axis}[%
width=0.95092\figurewidth,
height=\figureheight,
at={(0\figurewidth,0\figureheight)},
scale only axis,
xmin=1,
xmax=50,
xlabel={Rank},
ymin=0,
ymax=1,
ylabel={Identification rate (\%)},
axis x line*=bottom,
axis y line*=left,
legend style={at={($ (0,1) + (+0.1cm,+0.1cm) $)},anchor=north west,align=left,legend cell align=left,draw=black},
xmajorgrids,
ymajorgrids,
grid style={dashed}
]
\addplot [color=blue,solid,line width=1.0pt]
  table[row sep=crcr]{%
1    0.1518987341772152\\
5    0.34810126582278483\\
10   0.47468354430379744\\
15   0.5474683544303798\\
20   0.5886075949367089\\
25   0.629746835443038\\
30   0.6613924050632911\\
35   0.680379746835443\\
40   0.7120253164556962\\
45   0.7183544303797469\\
50   0.7310126582278481\\
};
\addlegendentry{DML};

\addplot [color=cyan,solid,line width=1.0pt]
  table[row sep=crcr]{%
1    0.14556962025316456\\
5    0.3322784810126582\\
10   0.4873417721518987\\
15   0.5569620253164557\\
20   0.5917721518987342\\
25   0.6424050632911392\\
30   0.689873417721519\\
35   0.7025316455696202\\
40   0.7278481012658228\\
45   0.7531645569620253\\
50   0.7848101265822784\\
};
\addlegendentry{DML, adaptation};

\end{axis}
\end{tikzpicture}%
        \centering\small{(a) Whole CUHK $\rightarrow$ VIPeR}
        \label{fig:allcuhk_viper}
  
    &
        \setlength\figureheight{3.5cm}
        \setlength\figurewidth{3.5cm}
        \begin{tikzpicture}[ampersand replacement=\&,font=\scriptsize]

\begin{axis}[%
width=0.95092\figurewidth,
height=\figureheight,
at={(0\figurewidth,0\figureheight)},
scale only axis,
xmin=1,
xmax=50,
xlabel={Rank},
ymin=0,
ymax=1,
ylabel={Identification rate (\%)},
axis x line*=bottom,
axis y line*=left,
legend style={at={($ (0,1) + (+0.1cm,+0.1cm) $)},anchor=north west,align=left,legend cell align=left,draw=black},
xmajorgrids,
ymajorgrids,
grid style={dashed}
]
\addplot [color=blue,solid,line width=1.0pt]
  table[row sep=crcr]{%
1      0.126582278481 \\
5      0.303797468354 \\
10      0.414556962025 \\
15      0.490506329114 \\
20      0.550632911392 \\
25      0.613924050633 \\
30      0.651898734177 \\
35      0.686708860759 \\
40      0.708860759494 \\
45      0.75 \\
50      0.759493670886 \\
};
\addlegendentry{DML};

\addplot [color=cyan,solid,line width=1.0pt]
  table[row sep=crcr]{%
1      0.120253164557 \\
5      0.29746835443 \\
10      0.462025316456 \\
15      0.541139240506 \\
20      0.594936708861 \\
25      0.617088607595 \\
30      0.645569620253 \\
35      0.667721518987 \\
40      0.696202531646 \\
45      0.712025316456 \\
50      0.73417721519 \\
};
\addlegendentry{DML, adaptation};

\end{axis}
\end{tikzpicture}%
        \centering\small{(b) CUHK/p1 $\rightarrow$ VIPeR}
    &
        \setlength\figureheight{3.5cm}
        \setlength\figurewidth{3.5cm}
        \begin{tikzpicture}[ampersand replacement=\&,font=\scriptsize]

\begin{axis}[%
width=0.95092\figurewidth,
height=\figureheight,
at={(0\figurewidth,0\figureheight)},
scale only axis,
xmin=1,
xmax=50,
xlabel={Rank},
ymin=0,
ymax=1,
ylabel={Identification rate (\%)},
axis x line*=bottom,
axis y line*=left,
legend style={at={($ (0,1) + (+0.1cm,+0.1cm) $)},anchor=north west,align=left,legend cell align=left,draw=black},
xmajorgrids,
ymajorgrids,
grid style={dashed}
]
\addplot [color=blue,solid,line width=1.0pt]
  table[row sep=crcr]{%
1      0.0664556962025 \\
5      0.167721518987 \\
10      0.253164556962 \\
15      0.275316455696 \\
20      0.316455696203 \\
25      0.348101265823 \\
30      0.379746835443 \\
35      0.414556962025 \\
40      0.45253164557 \\
45      0.474683544304 \\
50      0.496835443038 \\
};
\addlegendentry{DML};

\addplot [color=cyan,solid,line width=1.0pt]
  table[row sep=crcr]{%
1      0.0632911392405 \\
5      0.161392405063 \\
10      0.259493670886 \\
15      0.338607594937 \\
20      0.389240506329 \\
25      0.417721518987 \\
30      0.439873417722 \\
35      0.471518987342 \\
40      0.496835443038 \\
45      0.518987341772 \\
50      0.53164556962 \\
};
\addlegendentry{DML, adaptation};

\end{axis}
\end{tikzpicture}%
        \centering\small{(c) PRID $\rightarrow$ VIPeR}
 \end{tabular}

 \begin{tabular}{ p{5cm}  p{5cm}  p{5cm} }
        \setlength\figureheight{3.5cm}
        \setlength\figurewidth{4cm}
        \begin{tikzpicture}[ampersand replacement=\&,font=\scriptsize]

\begin{axis}[%
width=0.95092\figurewidth,
height=\figureheight,
at={(0\figurewidth,0\figureheight)},
scale only axis,
xmin=1,
xmax=50,
xlabel={Rank},
ymin=0,
ymax=1,
ylabel={Identification rate (\%)},
axis x line*=bottom,
axis y line*=left,
legend style={at={($ (0,1) + (+0.1cm,+0.1cm) $)},anchor=north west,align=left,legend cell align=left,draw=black},
xmajorgrids,
ymajorgrids,
grid style={dashed}
]
\addplot [color=blue,solid,line width=1.0pt]
  table[row sep=crcr]{%
1    0.08\\
5    0.13\\
10   0.22\\
15   0.27\\
20   0.31\\
25   0.35\\
30   0.4\\
35   0.41\\
40   0.41\\
45   0.43\\
50   0.44\\
};
\addlegendentry{DML};

\addplot [color=cyan,solid,line width=1.0pt]
  table[row sep=crcr]{%
1    0.07\\
5    0.19\\
10   0.27\\
15   0.32\\
20   0.35\\
25   0.37\\
30   0.39\\
35   0.41\\
40   0.43\\
45   0.45\\
50   0.45\\
};
\addlegendentry{DML, adaptation};

\end{axis}
\end{tikzpicture}%

        \centering\small{(d) Whole CUHK $\rightarrow$ PRID}
        \label{fig:allcuhk_prid}
    &
        \setlength\figureheight{3.5cm}
        \setlength\figurewidth{4cm}
        \begin{tikzpicture}[ampersand replacement=\&,font=\scriptsize]

\begin{axis}[%
width=0.95092\figurewidth,
height=\figureheight,
at={(0\figurewidth,0\figureheight)},
scale only axis,
xmin=1,
xmax=50,
xlabel={Rank},
ymin=0,
ymax=1,
ylabel={Identification rate (\%)},
axis x line*=bottom,
axis y line*=left,
legend style={at={($ (0,1) + (+0.1cm,+0.1cm) $)},anchor=north west,align=left,legend cell align=left,draw=black},
xmajorgrids,
ymajorgrids,
grid style={dashed}
]
\addplot [color=blue,solid,line width=1.0pt]
  table[row sep=crcr]{%
1      0.04 \\
5      0.08 \\
10      0.15 \\
15      0.22 \\
20      0.25 \\
25      0.3 \\
30      0.32 \\
35      0.36 \\
40      0.39 \\
45      0.41 \\
50      0.44 \\
};
\addlegendentry{DML};

\addplot [color=cyan,solid,line width=1.0pt]
  table[row sep=crcr]{%
1      0.06 \\
5      0.16 \\
10      0.21 \\
15      0.27 \\
20      0.31 \\
25      0.36 \\
30      0.39 \\
35      0.41 \\
40      0.41 \\
45      0.42 \\
50      0.43 \\
};
\addlegendentry{DML, adaptation};

\end{axis}
\end{tikzpicture}%
        \centering\small{(e) CUHK/p1 $\rightarrow$ PRID}
        \label{fig:cuhk_p1_prid}
    &
	       \setlength\figureheight{3.5cm}
       \setlength\figurewidth{4cm}
       \begin{tikzpicture}[ampersand replacement=\&,font=\scriptsize]

\begin{axis}[%
width=0.95092\figurewidth,
height=\figureheight,
at={(0\figurewidth,0\figureheight)},
scale only axis,
xmin=1,
xmax=50,
xlabel={Rank},
ymin=0,
ymax=1,
ylabel={Identification rate (\%)},
axis x line*=bottom,
axis y line*=left,
legend style={at={($ (0,1) + (+0.1cm,+0.1cm) $)},anchor=north west,align=left,legend cell align=left,draw=black},
xmajorgrids,
ymajorgrids,
grid style={dashed}
]
\addplot [color=blue,solid,line width=1.0pt]
  table[row sep=crcr]{%
1      0.08 \\
5      0.15 \\
10      0.19 \\
15      0.25 \\
20      0.28 \\
25      0.34 \\
30      0.35 \\
35      0.36 \\
40      0.39 \\
45      0.4 \\
50      0.41 \\
};
\addlegendentry{DML};

\addplot [color=cyan,solid,line width=1.0pt]
  table[row sep=crcr]{%
1      0.07 \\
5      0.19 \\
10      0.25 \\
15      0.27 \\
20      0.31 \\
25      0.36 \\
30      0.39 \\
35      0.42 \\
40      0.42 \\
45      0.46 \\
50      0.47 \\
};
\addlegendentry{DML, adaptation};

\end{axis}
\end{tikzpicture}%
        \centering\small{(f) VIPeR $\rightarrow$ PRID}
        \label{fig:viper_prid}  \\    
 \end{tabular}

 \begin{tabular}{ p{5cm}  p{5cm}}
        \setlength\figureheight{3.5cm}
        \setlength\figurewidth{4cm}
        \begin{tikzpicture}[ampersand replacement=\&,font=\scriptsize]

\begin{axis}[%
width=0.95092\figurewidth,
height=\figureheight,
at={(0\figurewidth,0\figureheight)},
scale only axis,
xmin=1,
xmax=50,
xlabel={Rank},
ymin=0,
ymax=1,
ylabel={Identification rate (\%)},
axis x line*=bottom,
axis y line*=left,
legend style={at={($ (0,1) + (+0.1cm,+0.1cm) $)},anchor=north west,align=left,legend cell align=left,draw=black},
xmajorgrids,
ymajorgrids,
grid style={dashed}
]
\addplot [color=blue,solid,line width=1.0pt]
  table[row sep=crcr]{%
1      0.109053497942 \\
5      0.265432098765 \\
10      0.366255144033 \\
15      0.407407407407 \\
20      0.465020576132 \\
25      0.491769547325 \\
30      0.510288065844 \\
35      0.541152263374 \\
40      0.572016460905 \\
45      0.59670781893 \\
50      0.617283950617 \\
};
\addlegendentry{DML};

\addplot [color=cyan,solid,line width=1.0pt]
  table[row sep=crcr]{%
1      0.125514403292 \\
5      0.253086419753 \\
10      0.366255144033 \\
15      0.440329218107 \\
20      0.5 \\
25      0.5329218107 \\
30      0.572016460905 \\
35      0.594650205761 \\
40      0.617283950617 \\
45      0.650205761317 \\
50      0.668724279835 \\
};
\addlegendentry{DML, adaptation};

\end{axis}
\end{tikzpicture}%
        \centering\small{(g) VIPeR $\rightarrow$ CUHK/p1}
        \label{fig:viper_cuhk_p1}
    &
        \setlength\figureheight{3.5cm}
        \setlength\figurewidth{4cm}
        \begin{tikzpicture}[ampersand replacement=\&,font=\scriptsize]

\begin{axis}[%
width=0.95092\figurewidth,
height=\figureheight,
at={(0\figurewidth,0\figureheight)},
scale only axis,
xmin=1,
xmax=50,
xlabel={Rank},
ymin=0,
ymax=1,
ylabel={Identification rate (\%)},
axis x line*=bottom,
axis y line*=left,
legend style={at={($ (0,1) + (+0.1cm,+0.1cm) $)},anchor=north west,align=left,legend cell align=left,draw=black},
xmajorgrids,
ymajorgrids,
grid style={dashed}
]
\addplot [color=blue,solid,line width=1.0pt]
  table[row sep=crcr]{%
1      0.0569620253165 \\
5      0.139240506329 \\
10      0.212025316456 \\
15      0.284810126582 \\
20      0.316455696203 \\
25      0.344936708861 \\
30      0.370253164557 \\
35      0.389240506329 \\
40      0.414556962025 \\
45      0.443037974684 \\
50      0.465189873418 \\
};
\addlegendentry{DML};

\addplot [color=cyan,solid,line width=1.0pt]
  table[row sep=crcr]{%
1      0.0843621399177 \\
5      0.191358024691 \\
10      0.263374485597 \\
15      0.316872427984 \\
20      0.347736625514 \\
25      0.395061728395 \\
30      0.432098765432 \\
35      0.471193415638 \\
40      0.504115226337 \\
45      0.5329218107 \\
50      0.553497942387 \\
};
\addlegendentry{DML, adaptation};

\end{axis}
\end{tikzpicture}%
        \centering\small{(h) PRID $\rightarrow$ CUHK/p1}
        \label{fig:prid_cuhk_p1}
  \\    
  \end{tabular}
  \caption{Results on VIPeR, PRID and CUHK/p1 with and without domain-adversarial learning. Across the eight domain pairs domain-adversarial learning improves re-identification accuracy. For some domain pairs the improvement is considerable.}
  \label{fig:adaptresults}
\end{figure}
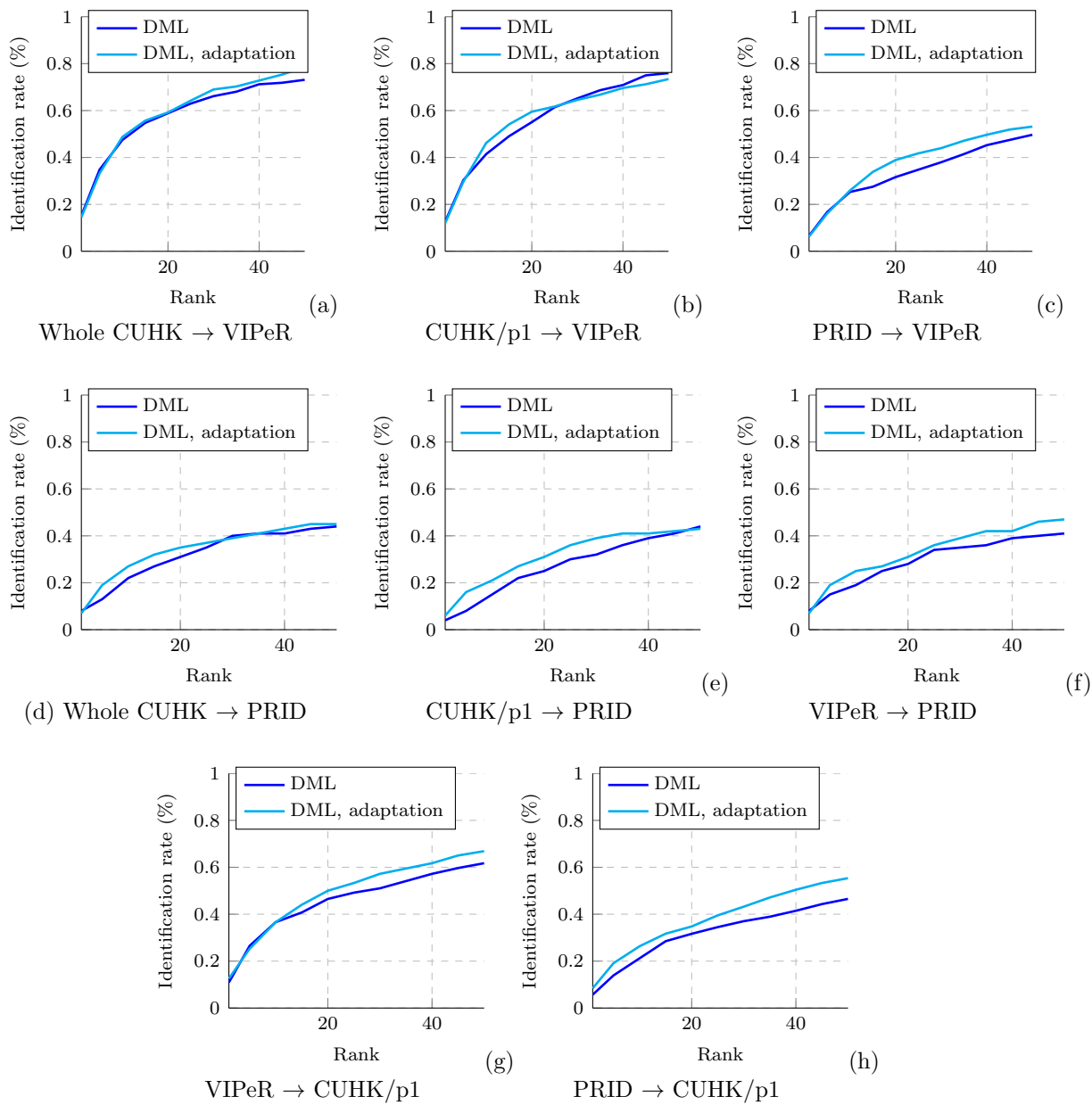
\clearpage

\begin{figure*}
\centering
\begin{tabular}{c c}
\includegraphics[height=7cm,width=7cm]{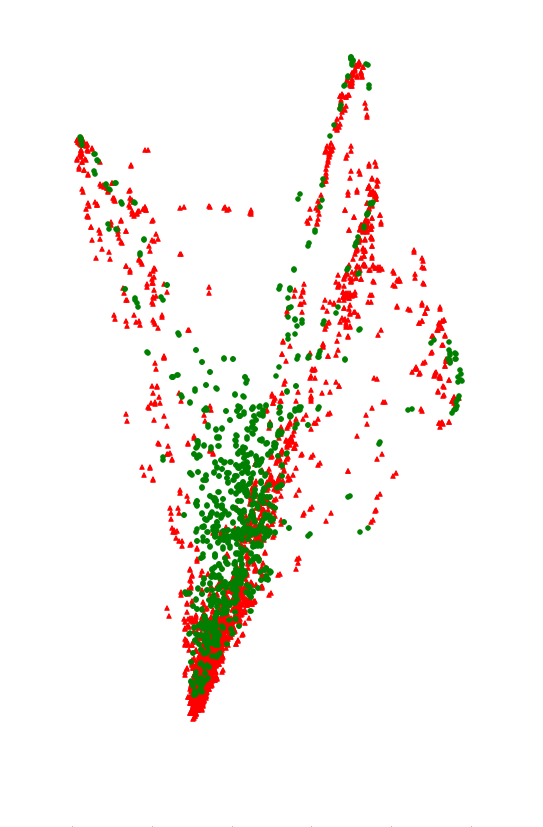}&
\includegraphics[height=7cm,width=7cm]{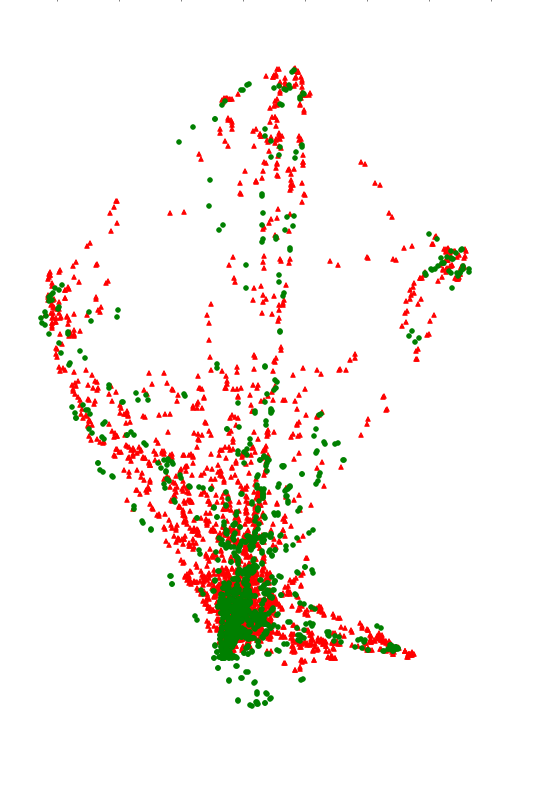}\\ 
\small{(a) DML}&
\small{(b) DML, adaptation}\\
\end{tabular}
\caption{The effect of adaptation shown by t-SNE visualizations of source and target domains descriptors in a VIPeR $\rightarrow$ CUHK/p1 experiment pair. VIPeR is depicted with \textit{green} and CUHK/p1 - with \textit{red}. As in the image classification case, domain-adversarial learning ensures a closer match between the source and the target distributions. }
\label{fig:reidtsne}
\end{figure*}

}

After the sufficient number of iterations, domain-adversarial training consistently improves the performance of re-identification. For the pairs that involve PRID data set, which is more dissimilar to the other two data sets, the improvement is considerable. Overall, this demonstrates the applicability of the domain-adversarial learning beyond classification problems.

Figure \ref{fig:reidtsne} further demonstrates the effect of adaptation on the distributions of the learned descriptors in the source and in target sets in VIPeR $\rightarrow$ CUHK/p1 experiments, where domain adversarial learning once again achieves better intermixing of the two domains.

\section{Conclusion}

The paper proposes a new approach to domain adaptation of feed-forward neural networks, which allows large-scale training based on large amount of annotated data in the source domain and large amount of unannotated data in the target domain. Similarly to many previous shallow and deep DA techniques, the adaptation is achieved through aligning the distributions of features across the two domains. However, unlike previous approaches, the alignment is accomplished through standard backpropagation training.

The approach is motivated and supported by the domain adaptation theory of \citet{BenDavid-NIPS06,BenDavid-MLJ2010}. 
The main idea behind DANN is to enjoin the network hidden layer to learn a representation which is predictive of the source example labels, but uninformative about the domain of the input (source or target). 
We implement this new approach within both shallow and deep feed-forward architectures. The latter allows simple implementation within virtually any deep learning package through the introduction of a simple gradient reversal layer. 
We have shown that our approach is flexible and achieves state-of-the-art results on a variety of benchmark in domain adaptation, namely for sentiment analysis and image classification tasks. 

A convenient aspect of our approach is that the domain adaptation component can be added to almost any neural network architecture that is trainable with backpropagation. Towards this end, We have demonstrated experimentally that the approach is not confined to classification tasks but can be used in other feed-forward architectures, \eg, for descriptor learning for person re-identification.

\acks{This work has been supported by National Science and Engineering Research Council (NSERC) Discovery
grants 262067 and 0122405 as well as the Russian Ministry of Science and Education grant RFMEFI57914X0071.
Computations were performed on the Colosse supercomputer grid at Universit\'e Laval, under the auspices of Calcul Qu\'ebec and Compute Canada. The operations of Colosse are funded by the NSERC, the Canada Foundation for Innovation (CFI), NanoQu\'ebec, and the Fonds de recherche du Qu\'ebec -- Nature et technologies (FRQNT). We also thank the Graphics \& Media Lab, Faculty of Computational Mathematics and Cybernetics, Lomonosov Moscow State University for providing the synthetic road signs data set.}

\bibliography{ganin15a}

\end{document}